\documentclass{article}

 \usepackage[preprint]{neurips_2026}

\usepackage[utf8]{inputenc} %
\usepackage[T1]{fontenc}    %
\usepackage{hyperref}       %
\usepackage{url}            %
\usepackage{booktabs}       %
\usepackage{amsfonts}       %
\usepackage{nicefrac}       %
\usepackage{microtype}      %
\usepackage{xcolor}         %
\usepackage{amsmath}
\usepackage{enumitem}
\usepackage{multirow}
\usepackage{graphicx}
\usepackage{float}
\usepackage{longtable}
\usepackage[most]{tcolorbox}
\usepackage{color, soul}
\usepackage{booktabs}
\usepackage{tabularx}
\usepackage{array}
\usepackage{xspace}
\usepackage{tcolorbox}
\usepackage{pgfplots}
\pgfplotsset{compat=1.18}
\usepackage[capitalise]{cleveref}
\newcolumntype{Y}{>{\raggedright\arraybackslash}X}
\newcommand{\scorecell}[2]{\ensuremath{#1 \rightarrow #2}}

\usepackage{xcolor,booktabs,multirow,graphicx}

\usepackage{colortbl}
\usepackage{array}
\newtcolorbox{prompttemplatebox}[1]{
  colback=white!3!white,
  colframe=white!60!gray,
  title=#1,
  fonttitle=\bfseries\sffamily\small,
  arc=2pt,
  boxrule=0.55pt,
  enhanced jigsaw,
  fontupper=\ttfamily\footnotesize,
  left=4pt, right=4pt, top=2pt, bottom=2pt,
}
\newtcolorbox{promptexamplebox}[1]{
  colback=gray!5!white,
  colframe=gray!55!black,
  title=#1,
  fonttitle=\bfseries\sffamily\small,
  arc=2pt,
  boxrule=0.55pt,
  enhanced jigsaw,
  fontupper=\ttfamily\footnotesize,
  left=4pt, right=4pt, top=2pt, bottom=2pt,
}
\usepackage{wrapfig}
\newcommand{\dlift}{\ensuremath{\Delta_{\text{lift}}}}
\newcommand{\wav}{\texttt{\textsc{WavEHR}}\xspace}

\usepackage{xcolor,booktabs,graphicx}
\newcommand{\diagbin}[3]{\ensuremath{#1\,/\,#2\,/\,#3}}
\newcommand{\diagmulti}[2]{\ensuremath{#1\,/\,#2}}
\usepackage{xcolor,booktabs,multirow,graphicx}
\usepackage{subcaption} %

\newcommand{\subcell}[3]{%
  \ensuremath{#1_{\textcolor{gray}{\scriptscriptstyle[#2,#3]}}}%
}

\usepackage[most]{tcolorbox}

\usepackage[most]{tcolorbox}

\usepackage[most]{tcolorbox}

\newtcolorbox{takeaway}[1]{
    enhanced,
    breakable,
    colback=black!6,
    colframe=black!85,
    colbacktitle=black!85,
    coltitle=white,
    fonttitle=\normalsize\bfseries,
    fontupper=\normalsize,
    title={#1},
    boxrule=0.7pt,
    arc=2.5pt,
    outer arc=2.5pt,
    boxsep=1pt,
    left=4pt,
    right=4pt,
    top=5pt,
    bottom=3pt,
    before skip=4pt,
    after skip=4pt,
    attach boxed title to top left={
        xshift=5pt,
        yshift=-1.5mm
    },
    boxed title style={
        boxrule=0pt,
        arc=2.5pt,
        boxsep=1pt,
        left=4pt,
        right=4pt,
        top=1pt,
        bottom=1pt
    }
}

\definecolor{ao(english)}{rgb}{0.0, 0.5, 0.0}
\definecolor{awesome}{rgb}{1.0, 0.13, 0.32}

\usepackage{array,colortbl}
\usepackage{arydshln}

\usepackage{graphicx,booktabs, multirow, caption}

\title{How Well Do Clinical Foundation Models Leverage Structured Health Data and Waveforms?}

\author{%
 \textbf{Megha Thukral}\textsuperscript{1}\,\thanks{%
  Corresponding author: \texttt{mthukral3@gatech.edu}.} \quad
  \textbf{Yusuf Ali}\textsuperscript{1} \quad
  \textbf{Harish Haresamudram}\textsuperscript{2} \quad
  \textbf{Shruthi K. Hiremath}\textsuperscript{3} \\[2pt]
  \textbf{Katrin H\"ansel}\textsuperscript{3} \quad
  \textbf{Zsolt Kira}\textsuperscript{1} \quad
  \textbf{Thomas Pl\"otz}\textsuperscript{1} \\[6pt]
  \textnormal{\textsuperscript{1}Georgia Institute of Technology} \\
  \textnormal{\textsuperscript{2}University of Illinois Urbana-Champaign} \\
  \textnormal{\textsuperscript{3}Optum AI},
  \textnormal{United States}
}
\begin{document}

\maketitle
        
\begin{abstract}
Multimodal Medical Large Language Models (MedLLMs) are increasingly being explored in various healthcare settings. 
So far, existing efforts have predominantly focused on integrating structured electronic health records (EHRs) with medical imaging and text data.
However, relatively little is known about the capability of such models to jointly reason over paired EHR data and physiological waveforms.
This is particularly important given the broad real-world usage and availability of waveforms for diagnosis, monitoring, and risk assessment.
In this work \wav, we perform a systematic evaluation of open-source MedLLMs in this underexplored yet clinically relevant multimodal setting.
We leverage two large-scale, publicly available, de-identified patient datasets containing complementary ECG and PPG waveform modalities and examine \emph{conditional modality lift}, i.e., the change in task performance when waveform inputs are added to EHR-only baselines.
We find that the lift is \textbf{inconsistent}: open-source MedLLMs exhibit varied multimodal integration capabilities across EHR and physiological waveform data.
To understand how MedLLM performance is affected by multimodal input, we conduct counterfactual analyses to distinguish genuine reliance on physiological waveforms from the effects of mere modality presence.
Additionally, through layer-wise attention analysis and linear probing of intermediate representations, we examine how MedLLMs attend to and encode multimodal inputs.
We find that MedLLMs generally allocate more attention to EHR text inputs than to waveform tokens, and that waveform-directed attention does not consistently align with multimodal performance gains.
Finally, we investigate multimodal in-context learning as a training-free adaptation strategy and find that a few labeled waveform–EHR examples can improve multimodal performance in some model-task settings.
Overall, \wav provides a systematic characterization of MedLLMs' ability to jointly leverage EHR data and physiological waveforms, highlighting cross-modal integration limtations and identifying opportunities for advancing multimodal clinical modeling.
\end{abstract}

\section{Introduction}
Clinical decision-making often draws from multiple sources of patient information, including history, diagnoses, medications, laboratory trends, prior procedures, imaging, and physiological measurements~\cite{acosta2022multimodal,azarfar2025responsible}.
This has produced broader interest in multimodal medical AI systems that can combine heterogeneous clinical modalities when they provide complementary information \cite{rajpurkar2022ai,moor2023foundation,soenksen2022integrated}.
For example, recent Medical Large Language Models (MedLLMs) such as QoQ-Med \cite{qoqmed,qoqmed3} flexibly input any combination of electronic health records (EHRs), images including MRIs, ultrasounds, and waveforms such as ECG, to reason about health conditions and diagnoses.
Similarly, MedGemma~\cite{sellergren2025medgemma, sellergren2026medgemma} supports processing medical text and image understanding for multimodal clinical analysis.

In this work, we focus on a clinically important multimodal setting: EHR records paired with physiological signals (or waveforms).
This setting is important because EHRs and waveforms are broadly available across monitored care settings and can provide complementary information.
EHRs summarize longitudinal clinical context, while waveforms provide high-resolution physiological measurements that can support more informed clinical reasoning and prediction \cite{soenksen2022integrated,chen2023multimodal}.

Recent work has comprehensively evaluated large language models (LLMs) on EHR-based tasks, including their zero-shot generalisation to complex clinical prediction problems \cite{medhelm,yang2026ehrstruct,healthbench,medagentgym}.
In parallel, LLMs have demonstrated promising capabilities in interpreting physiological time-series signals \cite{li2026hearts,langer2025opentslm,healthllm,sensorlm,xu2026well}, and
foundation models have been developed for physiological and biosignals stemming from, e.g.,  ECG and PPG \cite{kim2024learning,pillai2024papagei}.
Yet, these research directions have largely evolved independently: studies on EHRs have primarily centred on textual and structured clinical reasoning, while analyzing 
physiological signals has focused on representation learning and modelling sensor data through language-based frameworks.
Emerging work \cite{qoqmed,langer2025opentslm} has started combining EHR data with physiological signals through modality-specific encoders \cite{kim2024learning}, but evaluations of existing Medical LLMs (MedLLMs) under this modality combination remain underexplored.
To this end, we conduct a systematic assessment to characterize if and how physiological waveforms, when combined with structured patient context, contribute to improved clinical performance in MedLLMs.

In this work, \wav, we perform evaluations of EHR-waveform multimodal tasks using leading open-source MedLLMs. 
We study this setting through two large-scale datasets that differ in care context and physiological signal type: 
\emph{(i) MIMIC-IV} with structured EHR and diagnostic 12-lead ECGs \cite{mimiciv,mimicivecg}; and \emph{(ii) MC-MED}, with Emergency Department's EHR with bedside ECG and PPG ~\cite{mcmed}. 
These settings cover both diagnostic recordings and bedside physiological monitoring.
Overall, \wav covers four \textit{zero-shot prediction} tasks spanning mortality, length of stay,  clinical deterioration, and ED revisit prediction.
The evaluation is performed on \textit{three leading 
MedLLMs}: MedGemma 4B, MedGemma 27B, and QoQ-Med \cite{sellergren2025medgemma,qoqmed}.

Across datasets and models, we measure the change in performance (\dlift) from adding physiological waveform data to fixed EHR-only inputs.
We quantify uncertainty using bootstrap confidence intervals and report results across demographic, clinical, and temporal stratifications.
From our zero-shot evals, we find that, in many cases, MedLLMs are able to leverage EHR the physiological signal modality to boost performance over the EHR-only evaluation.
However, we find that such $\dlift$ is  \textit{inconsistent}, varying across models, tasks, and subgroups, with heterogeneous performance being obtained across model-task settings.
Despite physiological waveforms carrying complementary information that can be vital to clinical decision-making, our findings highlight \textit{limitations in the ability of current MedLLMs to effectively integrate these modalities}, offering opportunities for future multimodal research in this important field.

Our contributions are threefold:
\begin{enumerate}
    \item We introduce \wav, a systematic evaluation of open-source MedLLMs in the underexplored setting of joint EHR and physiological waveform modeling, examining when waveform information improves clinical prediction and whether MedLLMs effectively leverage the added modality.

    \item We evaluate MedGemma-4B, MedGemma-27B, and QoQ-Med across four zero-shot clinical prediction tasks spanning two clinical settings: diagnostic ECG interpretation and emergency-department bedside monitoring. We find that conditional modality lift is inconsistent across models and tasks, highlighting limitations in cross-modal integration.

    \item We conduct counterfactual waveform experiments, layer-wise attention, and linear probing analyses to delineate how MedLLMs integrate physiological waveform and EHR data. We further investigate whether multimodal in-context learning can improve such cross-modal integration without additional training.
\end{enumerate}

\section{Related Work} 
\textbf{Medical MLLMs and Multimodal Clinical Evaluation. }
Recent advancements in multimodal foundation models have catalyzed the emergence of a burgeoning line of medical large language models (LLMs) that adapt general image-text and multimodal architectures to biomedical and clinical tasks. 
Early efforts such as Med-Flamingo~\cite{moor2023med}, LLaVA-Med~\cite{li2023llava} and CXR-LLaVA~\cite{lee2025cxr} helped establish medical image--text reasoning as a central evaluation setting, while more recent models including MedGemma~\cite{sellergren2025medgemma} and the QoQ-Med family~\cite{qoqmed,qoqmed3} broaden this paradigm, covering medical imaging, clinical text, and EHR, and, in the case of QoQ-Med, physiological signals.
Overall, multimodal medical LLMs (MedLLMs) have enabled the integration/analysis of heterogeneous data types, such as structured Electronic Health Records (EHRs) and physiological waveforms, thereby mirroring the comprehensive, multi-evidence approach used by clinicians to improve diagnostic accuracy, risk stratification, and patient monitoring.

Benchmarks have evolved in tandem with the advancing capabilities of these multimodal models.  
GMAI-MMBench~\cite{chen2024gmai}, OmniMedVQA~\cite{hu2024omnimedvqa}, and MedHELM~\cite{medhelm} assess broad medical capability across domains, modalities, and task types.
EHRXQA~\cite{ehrxqa}, RadFusion~\cite{zhou2021radfusion}, and INSPECT~\cite{huang2023inspect} bring structured EHR information into multimodal evaluation with medical images. 
A related line of physiological signal-centered systems, including PULSE-ECG~\cite{liu2024teach}, OpenTSLM~\cite{langer2025opentslm}, UniPACT~\cite{tang2026unipact}, GEM~\cite{lan2025gem}, HEARTS~\cite{li2026hearts}, and SignalMC-MED~\cite{gustafsson2026signalmc}, study ECGs, health time series, and biosignal representation learning. 
Overall, these works establish medical multimodality, structured-record reasoning, and physiological signals as essential evaluation regimes. 
Building on this, we perform zero-shot evaluation and benchmark the capability of leading MedLLMs to characterize their modality interactions and relative contributions of EHR and waveforms.

\textbf{Structured EHR and Physiological Signals. }
Structured-EHR benchmarks and evaluation works such as EHRSQL~\cite{lee2022ehrsql}, DrugEHRQA~\cite{bardhan2022drugehrqa}, ClinicRealm~\cite{zhu2026clinicrealm}, EHRStruct~\cite{yang2026ehrstruct}, and VirtualEHR~\cite{jiang2025medagentbench} show that clinical records pose distinct challenges such as schema grounding, temporal reasoning, long-context integration, and heterogeneous clinical variables. 
A parallel line of work targets physiological signals as a complementary view of patient state:
MC-BEC~\cite{chen2023multimodal} pairs emergency-care records with continuous waveforms and other clinical inputs, HEARTS~\cite{li2026hearts} evaluates health time-series-only reasoning in leading LLMs, and SignalMC-MED~\cite{gustafsson2026signalmc} focuses on benchmarking specialized biosignal foundation models on synchronized ECG--PPG biosignals. 
Together, these lines of work motivate the EHR--waveform setting, where structured records provide clinical context, and waveforms provide
physiological measurements.
In contrast, \wav targets assessment of medical MLLMs that can jointly leverage long-form patient visit data and physiological signals.
It does not assume ECG-specific architectures, prompting strategies, or task-specific signal fusion.

\textbf{Subgroup and Modality-Contribution Analysis. }
We examine how the addition of physiological waveforms contributes to performance from two complementary perspectives: 
\textit{i)} subgroup-wise performance; and 
\textit{ii)} attention-based analysis.
WILDS~\cite{koh2021wilds} formalizes domain and subpopulation shifts. MEDFAIR~\cite{zong2022medfair}, FairMedFM~\cite{jin2024fairmedfm}, and related fairness studies~\cite{ktena2024generative,jclinepi2024sociodem} further show that aggregate performance can obscure substantial variation across demographic and clinical strata. 
Prior work on MIMIC~\cite{meng2022interpretability} similarly evaluates clinical prediction along axes such as demographics, comorbidity burden, and age. Motivated by these studies, we analyze modality contribution across patient subgroups, asking whether the benefit of waveform data shifts across groups and clinical settings when the structured EHR context is fixed.
In addition, we analyze attention maps to understand how the model uses waveform information. This connects to prior work on modality contribution, reliability, and dominance in multimodal models, which shows that models may rely unevenly on available inputs or collapse toward a dominant modality~\cite{parcalabescu2023mm,wu2025mitigating,wei2025improving}. 
Accordingly, \wav analyses subgroup-stratified conditional lift and complements it with waveform-directed attention analysis.

\begin{figure}
    \centering
    \includegraphics[width=\linewidth]{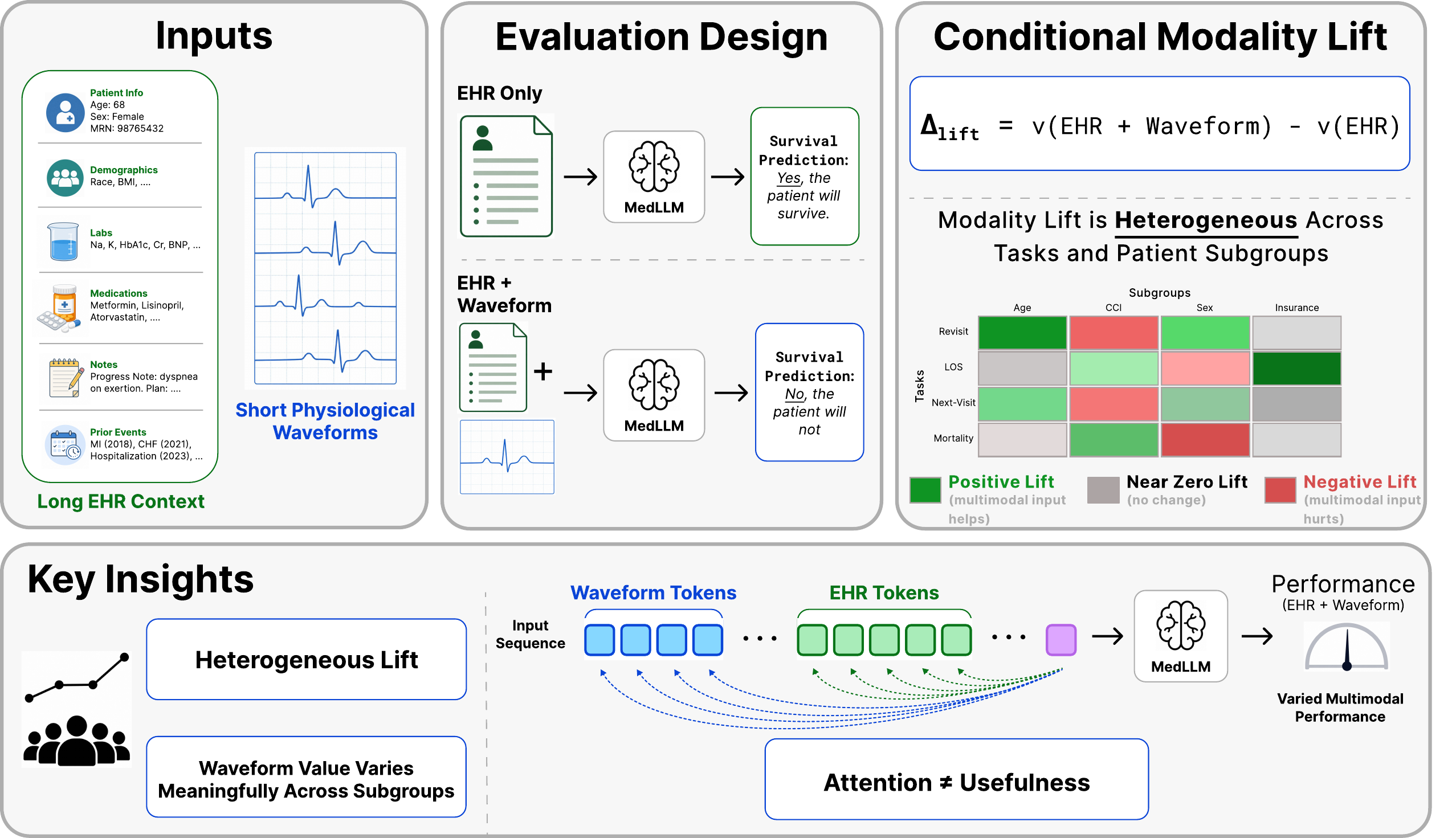}
    \caption{\wav: a framework for evaluating the capability of frontier, open-source medical LLMs to leverage physiological waveforms along with structured EHR data for clinical outcome predictions.
    }
    \label{fig:wavehr_framework}
\end{figure}
\section{Evaluation Framework}
\label{sec:exp}

Our evaluation framework is visualized in \cref{fig:wavehr_framework}.
The MedLLMs receive two types of input: \textit{(i)} textualized structured EHR narrative containing patient demographics, history, etc; and 
\textit{(ii)} short duration physiological waveform segments (e.g., 12-lead ECG).
First, we pass only textualized EHR to the MedLLM, to solve the task, e.g., deterioration prediction.
Subsequently, we pass both EHR and waveform signals (e.g., 12-lead ECG).
The conditional modality lift ($\dlift$) is defined as the difference in task performance between the multimodal and EHR-only setting.
We conduct both aggregate and patient subgroup analyses to assess whether the performance gains from incorporating waveform data are consistent across models, tasks and populations.

We organize this analysis around four research questions that progressively examine multimodal model behavior. 
First, we evaluate whether multimodal EHR and waveform inputs improve \textit{zero-shot} MedLLM performance compared with EHR-only inputs across models, tasks, and datasets (\textbf{RQ1}).
We then ask whether the gains or reductions are \textit{consistent} across subgroups, since overall performance can mask uneven impacts across cohorts (\textbf{RQ2}). 
Next, we quantify the attention allocated to waveform tokens alongside long-context EHR tokens and use layer-wise linear probes to assess whether EHR and waveform representations contain task-relevant information and how their discriminability evolves across model layers (\textbf{RQ3}).
Finally, we explore whether multimodal in-context learning can improve model performance (\textbf{RQ4}).

\subsection{Datasets and Physiological Modalities}
\label{sec:data}

We evaluate on two large-scale clinical datasets that provide complementary and clinically diverse test settings: 
\textit{(i)}  MIMIC-IV \cite{mimiciv} covers inpatient  care, incl.\ ward and ICU admissions, and its structured EHRs are paired with diagnostic 12-lead ECG (from  MIMIC-IV-ECG \cite{mimicivecg}); and 
\textit{(ii) }MC-MED~\cite{mcmed} contains emergency-department care with continuous bedside-monitor waveforms, incl.\ ECG and PPG. 
Together, they enable evaluation across meaningful distribution shifts in cohort composition, clinical workflow, acquisition modality, and temporal context.
We provide a summary in \Cref{tab:data}. For both datasets, structured EHR narratives are textualized and sent to MedLLM as long-context patient input.
Physiological signals are processed using model-specific input formats: for MedGemma, signals are rendered as images, whereas for QoQ-Med they are provided as time-series inputs. 
For MIMIC-IV, waveform segments are rendered as a multi-panel 4$\times$2 ECG layout, similar to existing image-based ECG methods \cite{liu2024teach} unlike for MC-MED where it is renderd as one lead image.
We provide further data preparation details in App. \cref{sec:ehr_text_construction}.

\begin{table}[h]
\centering
\small
\begin{tabular}{@{}p{2.2cm}p{2.3cm}p{3.6cm}p{4.5cm}@{}}
\toprule
\textbf{Dataset} & \textbf{Setting} & \textbf{Physiological modality} & \textbf{Tasks} \\
\midrule
MIMIC-IV & Inpatient hospital admissions & 12-lead diagnostic ECG & 48-hour in-hospital mortality (\texttt{48-IHM}); length-of-stay (\texttt{LOS}) \\
\addlinespace
MC-MED & Emergency department &  Bedside ECG lead II; PPG/Pleth &  ED-Decompensation (\texttt{ED-Decomp}); ED revisit (\texttt{ED-Revisit}) \\
\bottomrule
\end{tabular}
\vspace{1mm}
\caption{Datasets, clinical settings, physiological modalities, and evaluated tasks.
}
\label{tab:data}
\end{table}

\subsection{Tasks}
\label{sec:tasks}

We utilize four prediction tasks in total to evaluate whether physiological waveforms add predictive value beyond longitudinal EHR context in inpatient and emergency care.
On MIMIC-IV, we evaluate \emph{(i)} 48-hour in-hospital mortality (\texttt{48-IHM}) and \emph{(ii)} length of stay with diagnostic ECGs (\texttt{LOS}), drawing on CLIMB and QoQ-Med~\cite{dai2025climb,qoqmed}.
On MC-MED, we adapt two tasks from MC-BEC~\cite{chen2023multimodal}: \emph{(iii)} near-term ED decompensation and \emph{(iv)} post-discharge ED revisit, evaluating bedside ECG and PPG. For each task, we compare EHR-only and multimodal EHR with waveform predictions on the same cases.
We evaluate supervised fusion baselines as a reference to contextualise the incremental value provided by the waveform signal on the respective task.
We provide more details on task prediction cutoffs and label construction in App. \Cref{app:tasks}.

\subsection{Models}
\label{sec:models}

We evaluate two state-of-the-art MedLLM model families: QoQ-Med \cite{qoqmed} and MedGemma \cite{sellergren2025medgemma}.
QoQ-Med is a Qwen-based clinical multimodal reasoning model designed to jointly process clinical images, physiological time-series, and text.
Its architecture prepends an ECG foundation model time-series encoder (ECG-JEPA~\cite{kim2024learning}) to a vision-language backbone, projects time-series outputs into the language-model token space, and interleaves projected image patches, time-series tokens, and tokenized text at inference. 
Thus,  QoQ-Med serves as our representative model for native physiological time-series ingestion within an integrated biosignal foundation model.

MedGemma \cite{sellergren2025medgemma} is a general-purpose medical vision-language model optimized for medical image and text comprehension based on the Gemma-3 architecture \cite{kamath2025gemma}.
Its multimodal variants utilize a SigLIP image encoder and accept image-text inputs. 
Because MedGemma does not natively ingest raw ECG/PPG time series, we evaluate it through its vision-language encoder by rendering ECG and PPG signals as grid-style waveform plots. 
We include both MedGemma-4B and MedGemma-27B to assess how model scaling alters performance.

\subsection{Supervised Baselines}
\label{sec:supervised_baselines}

To contextualize zero-shot MedLLM performance, we evaluate supervised EHR-only, waveform-only, and fusion baselines using the same cohorts, data splits, prediction time points, and waveform modalities. We extract features using frozen ClinicalBERT~\cite{huang2019clinicalbert} for EHR text, ECG-JEPA~\cite{kim2024learning} for ECG, and PaPaGei~\cite{pillai2024papagei} for PPG. We train task-specific linear classifiers on EHR features, waveform features, or their concatenation, and also evaluate late fusion of unimodal predictions. Details and results are provided in \Cref{app:supervised_baselines}.

\subsection{Zero-Shot Performance Analysis}
\label{sec:analysis_metrics}

\paragraph{Conditional Modality Lift $\dlift$.}
\hspace*{-0.7em}For an added physiological waveform  $s \in \{\mathrm{ECG}, \mathrm{PPG}\}$, we define
\begin{equation}
    \dlift^s = v(\mathrm{EHR}+s) - v(\mathrm{EHR}),
\end{equation}
where $v(\cdot)$ is the evaluation metric computed over the full set. 

For a subgroup $g$, we analogously define the subgroup-specific lift as
\begin{equation}
    \dlift^s(g) = v(\mathrm{EHR}+s;\,g) - v(\mathrm{EHR};\,g),
\end{equation}
where $v(\cdot;g)$ is computed only over samples belonging to subgroup $g$. Subgroups are derived from  metadata (as in previous work \cite{chen2023multimodal,meng2022interpretability}) -- and are not necessarily mutually exclusive. 
Unless otherwise stated, we report macro-F1 scores with 95\% confidence intervals using 1,000 bootstrap resamples.

\paragraph{Content-Normalized Attention Share.}

We adopt attention-based analysis from the Large Vision-Language Model (LVLM) literature, which aggregates answer-token attention over groups of input tokens to study modality dominance and collapse~\citep{zheng2025mllms,sim2025can,wu2025languageoverrules,park2025assessing}.
We extend this lens to clinical multimodal inputs by partitioning patient-content tokens into EHR and physiological-signal spans.
For each multimodal input, let $A_{\mathrm{EHR}}$ and $A_{wav}$ denote the self-attention weights assigned to the EHR-token span and the added waveform span, respectively. 
Following prior work ~\cite{wang2022interpretability, wang2023label}, these are measured at the final prefill query position, i.e., the state used to predict the first answer token, and averaged over heads and the included attention layers.

\begin{equation}
\mathrm{Attn}_{wav}
=
\frac{A_{wav}}{A_{wav} + A_{\mathrm{EHR}}}.
\end{equation}

This metric quantifies how attention over patient-content tokens is split between EHR and waveforms. 
It measures the fraction allocated to physiological-signal tokens rather than to EHR tokens. 
We exclude system, scaffold, and special-token positions from the denominator as decoder-style LLMs/LVLMs perform substantial allocation to attention-sink tokens~\citep{xiao2024streaming,sun2024massive} (see App. \Cref{fig:attention-sink}).

\section{Results}
\label{sec:results}

\newcommand{\dup}[1]{\textsubscript{\textcolor{gray}{(}\textcolor{green!60!black}{+#1}\textcolor{gray}{)}}}
\newcommand{\ddn}[1]{\textsubscript{\textcolor{gray}{(}\textcolor{red!80!black}{$-$#1}\textcolor{gray}{)}}}

\begin{table*}[t]
\centering
\small
\setlength{\tabcolsep}{3pt}
\renewcommand{\arraystretch}{1.2}
\setlength{\aboverulesep}{0pt}
\setlength{\belowrulesep}{0pt}
\setlength{\abovetopsep}{0pt}
\setlength{\belowbottomsep}{0pt}
\setlength{\cmidrulesep}{0pt}
\resizebox{\textwidth}{!}{%
\begin{tabular}{@{}
lll
>{\arrayrulecolor{gray!65}}l:
*{2}{>{\footnotesize\rmfamily}c}:
>{\arrayrulecolor{black}\footnotesize\rmfamily}c
*{5}{>{\footnotesize\rmfamily}c}
@{}}
\toprule
\multirow{2}{*}{\textbf{Dataset}}
& \multirow{2}{*}{\textbf{Task}}
& \multirow{2}{*}{\textbf{Signal}}
& \multicolumn{1}{>{\arrayrulecolor{gray!65}}l:}{%
    \multirow{2}{*}{\textbf{Metric}}}
& \multicolumn{2}{>{\arrayrulecolor{gray!65}}c:}{\textbf{Supervised}}
& \multicolumn{2}{>{\arrayrulecolor{black}}c}{\textbf{QoQ-Med-7B}}
& \multicolumn{2}{c}{\textbf{MedGemma-4B}}
& \multicolumn{2}{c}{\textbf{MedGemma-27B}} \\
\cmidrule(lr){5-6}
\cmidrule(lr){7-8}
\cmidrule(lr){9-10}
\cmidrule(lr){11-12}
& & &
\multicolumn{1}{>{\arrayrulecolor{gray!65}}l:}{}
& \texttt{\textsc{EHR}} & +\texttt{\textsc{wav}}\textsubscript{(\dlift)}
& \texttt{\textsc{EHR}} & +\texttt{\textsc{wav}}\textsubscript{(\dlift)}
& \texttt{\textsc{EHR}} & +\texttt{\textsc{wav}}\textsubscript{(\dlift)}
& \texttt{\textsc{EHR}} & +\texttt{\textsc{wav}}\textsubscript{(\dlift)} \\
\midrule

\multirow{4}{*}{\texttt{MIMIC-IV}}
& \multirow{2}{*}{\texttt{48-IHM}}
& \multirow{2}{*}{\shortstack[l]{ECG}}
& $\mathrm{F1}_{\mathrm{macro}}$
& 53.1 & 59.1\dup{6.0}
& 42.46 & 48.57\dup{6.11}
& \textbf{53.77} & \underline{53.19}\ddn{0.58}
& 37.16 & 42.53\dup{5.37} \\
& & & Acc.
& 71.2 & 83.2\dup{12.0}
& 55.15 & \underline{68.14}\dup{12.99}
& \textbf{78.89} & 65.66\ddn{13.23}
& 40.68 & 50.28\dup{9.60} \\

\cmidrule(l){2-12}
& \multirow{2}{*}{\texttt{LOS}}
& \multirow{2}{*}{\shortstack[l]{ECG}}
& $\mathrm{F1}_{\mathrm{macro}}$
& 27.2 & 28.4\dup{1.2}
& \underline{25.19} & \textbf{26.43}\dup{1.24}
& 12.88 & 12.98\dup{0.10}
& 24.15 & 21.99\ddn{2.16} \\
& & & Acc.
& 30.2 & 31.2\dup{1.0}
& \underline{31.47} & \textbf{35.36}\dup{3.89}
& 28.95 & 26.07\ddn{2.88}
& 26.92 & 24.32\ddn{2.60} \\

\midrule
\multirow{8}{*}{\texttt{MC-MED}}
& \multirow{2}{*}{\shortstack[l]{\texttt{ED-}\\\texttt{Decomp}}}
& \multirow{2}{*}{\shortstack[l]{ECG}}
& $\mathrm{F1}_{\mathrm{macro}}$
& 50.1 & 55.0\dup{4.9}
& 45.53 & 45.60\dup{0.07}
& 48.73 & 47.94\ddn{0.79}
& \textbf{54.58} & \underline{54.35}\ddn{0.23} \\
& & & Acc.
& 59.1 & 67.5\dup{8.4}
& 83.58 & 83.83\dup{0.25}
& \textbf{84.17} & \underline{84.00}\ddn{0.17}
& 83.57 & 83.01\ddn{0.56} \\

\cmidrule(l){2-12}
& \multirow{2}{*}{\shortstack[l]{\texttt{ED-}\\\texttt{Decomp}}}
& \multirow{2}{*}{\shortstack[l]{PPG}}
& $\mathrm{F1}_{\mathrm{macro}}$
& 49.8 & 53.1\dup{3.3}
& 45.94 & 46.04\dup{0.10}
& 49.41 & 48.76\ddn{0.65}
& \underline{55.95} & \textbf{56.92}\dup{0.97} \\
& & & Acc.
& 60.6 & 64.1\dup{3.5}
& 84.92 & 85.31\dup{0.39}
& \textbf{85.80} & \underline{85.76}\ddn{0.04}
& 84.56 & 84.18\ddn{0.38} \\

\cmidrule(l){2-12}
& \multirow{2}{*}{\shortstack[l]{\texttt{ED-}\\\texttt{Revisit}}}
& \multirow{2}{*}{\shortstack[l]{ECG}}
& $\mathrm{F1}_{\mathrm{macro}}$
& 23.8 & 29.0\dup{5.2}
& 23.83 & 23.84\dup{0.01}
& 23.84 & 23.96\dup{0.12}
& \underline{25.02} & \textbf{25.98}\dup{0.96} \\
& & & Acc.
& 52.8 & 78.9\dup{26.1}
& \textbf{87.94} & \underline{87.61}\ddn{0.33}
& 74.06 & 87.58\dup{13.52}
& 85.79 & 85.31\ddn{0.48} \\

\cmidrule(l){2-12}
& \multirow{2}{*}{\shortstack[l]{\texttt{ED-}\\\texttt{Revisit}}}
& \multirow{2}{*}{\shortstack[l]{PPG}}
& $\mathrm{F1}_{\mathrm{macro}}$
& 21.1 & 24.1\dup{3.0}
& 23.83 & 23.80\ddn{0.03}
& 23.83 & 23.32\ddn{0.51}
& \underline{25.00} & \textbf{26.26}\dup{1.26} \\
& & & Acc.
& 40.2 & 53.3\dup{13.1}
& \textbf{87.93} & \underline{87.48}\ddn{0.45}
& 74.05 & 75.77\dup{1.72}
& 85.92 & 85.47\ddn{0.45} \\

\bottomrule
\end{tabular}%
}
\arrayrulecolor{black}
\vspace{-1mm}
\caption{Performance (\%) on four clinical prediction tasks in \wav.
We find that open-source MedLLMs exhibit \textit{varied} multimodal
integration capabilities across EHRs and physiological waveform data.
Supervised +\texttt{wav} reports the best-performing fusion baseline from \Cref{app:supervised_baselines}.
In each row, the \textbf{best} MedLLM performance value is bold, and the \underline{second-best} is underlined.
See \Cref{tab:cond-lift-table} for \dlift values with $95\%$ CIs using bootstrap resamples.}
\label{tab:agg-perf-and-lift}
\end{table*}

\subsection{RQ1: Zero-Shot MedLLM Performance Analysis}
\label{sec:rq1}
We first analyse the multimodal performance of different MedLLMs across tasks and compare them to supervised baselines in \Cref{subsec:dlift_rq}. We then conduct counterfactual analyses to understand the reliance of MedLLMs on physiological waveforms and test robustness against waveform input representation in \Cref{subsec:counter}.

\subsubsection{Conditional Modality Lift across Models, Tasks, and Datasets}
\label{subsec:dlift_rq}
\Cref{tab:agg-perf-and-lift} presents zero-shot evaluations of MedLLMs using EHR data and physiological signals, alongside supervised EHR-only and multimodal fusion results for reference.
We note that supervised models trained with EHR and waveform fusion yield performance gains across all evaluation metrics, with improvements of varying magnitude across tasks (see App. \Cref{app:supervised_baselines} for full results).
This suggests that waveform signals can provide useful additional information that can be leveraged to improve task performance.
However, similar performance gains using multimodal inputs are not observed uniformly across MedLLMs-task settings.
\textbf{MIMIC-IV:}
On MIMIC-IV tasks, zero-shot MedLLMs generally achieve lower macro-F1 than the corresponding supervised baselines in multimodal settings.
For example, on \texttt{48-IHM}, supervised fusion achieves a macro-F1 of $59.1\%$, compared with approximately $53\%$ for the best-performing MedGemma-4B MedLLM on this task.
A closer look at the individual model performance reveals that QoQ-Med shows a \textit{clear positive conditional modality lift} of $+6.11\%$ with the incorporation of ECG signals in addition to EHR text across all MIMIC tasks. 
MedGemma-4B remains mostly stable in macro-F1, with no consistent performance gains in the multimodal setting, despite comparatively strong EHR-only performance on \texttt{48-IHM}.
MedGemma-27B degrades when ECG is added for \texttt{LOS}, but shows a substantial improvement over its EHR-only counterpart on \texttt{48-IHM}.
On analysing its responses, we observed that the EHR-only model is biased to predict the negative class, and adding ECG reduces this tendency, which may partly explain the improvement.
One plausible explanation for generally stronger multimodal performance gains in QoQ-Med is that it incorporates an ECG-JEPA encoder trained and aligned with Qwen2.5-VL and undergoes RL post-training to incentivise reasoning over ECG signals and EHR text \cite{qoqmed}.
MedGemma instead processes a dense visual representation of the $8$-lead ECG through its vision encoder.
These differences in signal representation and modality-aligned training may explain varying multimodal gains and highlight challenges in transferring MedLLMs to modalities and prediction tasks beyond their training distribution.

\textbf{MC-MED:}
On MC-MED tasks, supervised fusion across waveform-task settings increases macro-F1 by 3–5\% and accuracy by 5–12\%.
For example, on \texttt{ED-Decomp}, adding ECG improves macro-F1 under supervised fusion by $+5\%$, but yields little change across the three zero-shot MedLLMs.
QoQ-Med, despite benefiting from ECG on MIMIC-IV, shows only marginal gains when waveforms are added (both ECG and PPG), with small negative deltas for PPG on ED-Revisit.
MedGemma-4B shows slight positive gains on \texttt{ED-Revisit} but small negative deltas on \texttt{ED-Decomp}.
MedGemma-27B shows some macro-F1 gains on \texttt{ED-Revisit} with both signals and improves \texttt{ED-Decomp} with PPG.
We posit that QoQ-Med's limited benefit may partly reflect a larger distribution gap: MC-MED provides single-lead ECG and PPG signals, which differ from the multi-lead ECG used in MIMIC-IV and the waveform representation expected by the QoQ-Med signal encoder.
In contrast, MedGemma models receive the bedside signal as a rendered single-lead image, which is visually simpler than the dense visual information of the multiple leads of MIMIC-IV ECG rendering.
This may make single-lead ECG or PPG easier to process as a visual pattern, even without a native waveform encoder.
\begin{table*}[t]
\centering
\small
\setlength{\tabcolsep}{3pt}

\begin{subtable}[t]{0.49\linewidth}
\centering
\begin{tabular}{@{}lrr@{}}
\toprule
Input & QoQ-Med & MedGemma \\
\midrule
\texttt{Matched}
& 48.57
& 53.10 \\
\texttt{Noised}
& 45.21\,{\scriptsize\textcolor{red!80}{($-3.36$)}}
& 51.01\,{\scriptsize\textcolor{red!60}{($-2.09$)}} \\
\texttt{Shuffled}
& 45.02\,{\scriptsize\textcolor{red!60}{($-3.55$)}}
& 52.23\,{\scriptsize\textcolor{red!60}{($-0.87$)}} \\
\texttt{Mismatched}
& 48.42\,{\scriptsize\textcolor{red!60}{($-0.15$)}}
& 53.20\,{\scriptsize\textcolor{black!60}{($+0.10$)}} \\
\bottomrule
\end{tabular}
\caption{ECG counterfactual controls on \texttt{48-IHM}.}
\label{tab:ecg_counterfactuals}
\end{subtable}\hfill
\begin{subtable}[t]{0.49\linewidth}
\centering
\begin{tabular}{@{}lcc@{}}
\toprule
\textbf{Perturbation} & \texttt{48-IHM} & \texttt{LOS} \\
\midrule
\texttt{White} bg ($4\times2$)
& $-2.0_{\textcolor{gray}{[-3.6,-0.4]}}$
& $-0.1_{\textcolor{gray}{[-0.9,+0.7]}}$ \\
\texttt{Red} grid ($4\times2$)
& $-0.6_{\textcolor{gray}{[-2.6,+1.3]}}$
& $+0.1_{\textcolor{gray}{[-0.6,+1.1]}}$ \\
\bottomrule
\end{tabular}
\caption{Rendering sensitivity for MedGemma-4B.}
\label{tab:rq4_ingestion}
\end{subtable}

\caption{ECG control and rendering sensitivity analyses on MIMIC-IV.
(a) Macro-F1 (\%); parentheses show control minus matched
(patient-specific) ECG, in percentage points.
(b) $\dlift$ in macro-F1 relative to EHR alone, with
95\% paired-bootstrap CIs in gray subscripts.}
\label{tab:ecg_sensitivity}
\end{table*}

\subsubsection{Sensitivity to Waveform Content and Representation}
\label{subsec:counter}
\paragraph{Counterfactual Analysis.}

To assess whether changes in multimodal performance reflect patient-specific physiological information or the effect of adding a ``waveform-shaped" input, we perform controlled input signal perturbations: specifically, replace the matched ECG with Gaussian noise (\texttt{Noised}), a temporally shuffled copy of the same ECG (\texttt{Shuffled}), or an ECG from another patient (\texttt{Mismatched}) in \cref{tab:ecg_counterfactuals}. 
We keep the EHR text fixed and preserve the input format and modality tokens across the waveform conditions. 
Noise and shuffling disrupt waveform structure, while the mismatched ECG preserves a real waveform but breaks its pairing with the patient's EHR.
For \texttt{48-IHM}, QoQ-Med's macro-F1 increases from $42.46\%$ with EHR alone to $45.21\%$ with noise and $45.02\%$ with shuffled ECG, indicating that some lift occurs even with corrupted inputs.
However, the matched ECG reaches $48.57\%$, exceeding both corrupted controls by more than three points, but scores only $0.15$ points above the mismatched ECG.
For MedGemma, adding the matched ECG lowers performance relative to EHR alone, but the drop is smaller than with noise or shuffling, and the matched and mismatched ECGs yield nearly the same score.
Together, these controls suggest that both models are sensitive to waveform perturbations on \texttt{48-IHM}, but do not establish a benefit from correct patient–ECG pairing.
Nevertheless, similar aggregate performance with mismatched ECGs does not rule out patient-specific effects on individual predictions, as randomly selected replacement ECGs may still share morphological characteristics with the original recordings \cite{jekova2018human,park2015heartsearcher}.

\paragraph{Signal rendering-mode sensitivity.}
We use a $4 \times 2$ diagnostic grid on a clinical-style red grid background to
render $12$-lead ECGs in our main MIMIC-IV experiments. To assess sensitivity
to this choice, we compare it with the same grid on a plain white background
for MedGemma-4B, holding the underlying ECG segment and layout fixed (see
Table~\ref{tab:rq4_ingestion}). On \texttt{48-IHM}, the white background lowers
macro-F1 by $2.0$ points relative to the default
clinical-paper render. However, on \texttt{LOS}, the two backgrounds are within noise.
These results suggest that multimodal lift can be
sensitive to how physiological signals are presented to image-based MedLLMs.

\begin{takeaway}{Takeaway}
Overall, \textbf{RQ1} suggests that MedLLMs show variable performance across heterogeneous signal acquisition conditions and input distributions, with inconsistent gains from multimodal inputs combining EHR text and physiological signals. These findings point to generalisation challenges for multimodal clinical foundation models.
Broader integration of generic time-series encoders, together with lightweight adaptation to distribution shifts, may help enable more reliable multimodal clinical reasoning.
\end{takeaway}

\subsection{RQ2: Subgroup Differences in Conditional Modality Lift}

\begin{figure}[t]
    \centering
     \includegraphics[width=\linewidth]{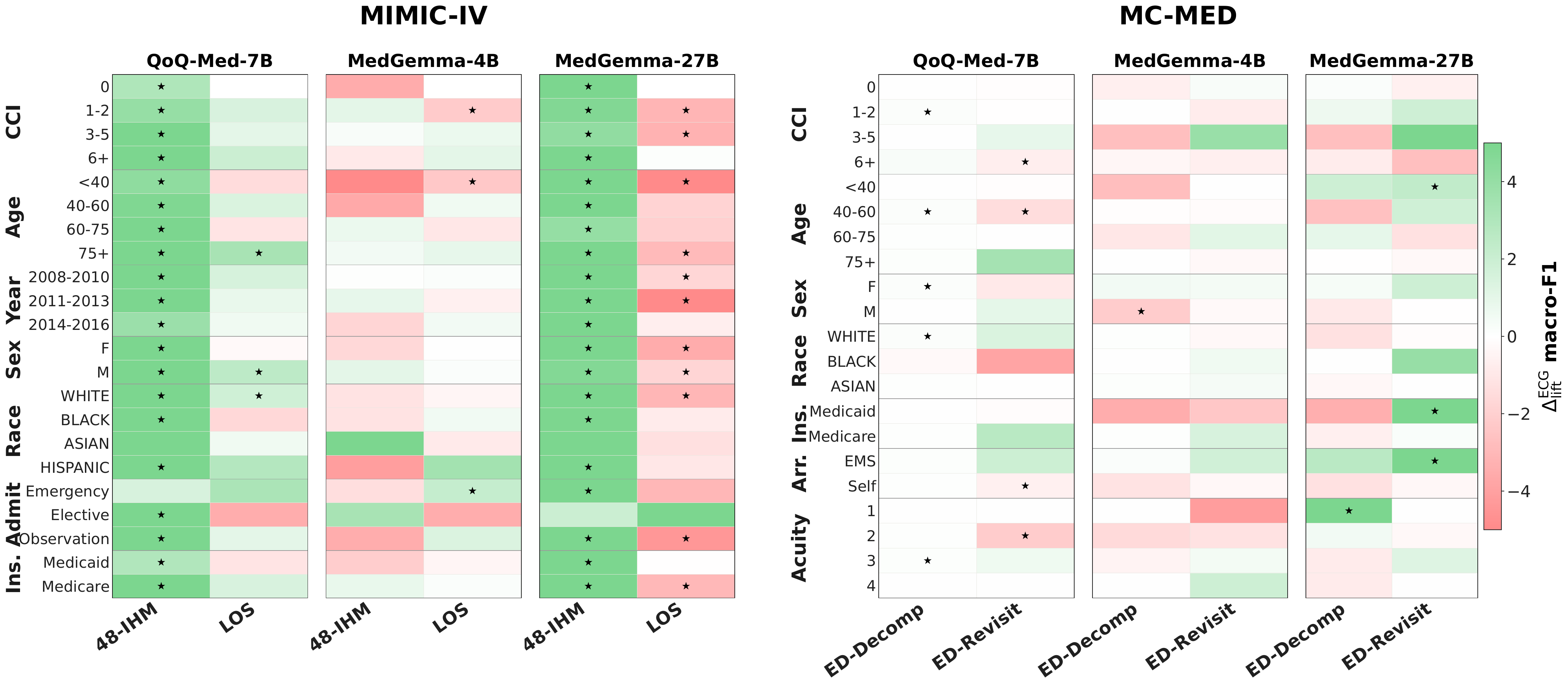}
    \caption{
   Subgroup-wise conditional modality lift ($\dlift$) across four prediction tasks in \wav. Green indicates positive $\dlift$, and red indicates negative $\dlift$. Exact values are reported in App.~\cref{app:subgroup_tables}. On MIMIC-IV, subgroups generally show gains when the full cohort shows gains (see \cref{tab:agg-perf-and-lift} for full cohort results) and declines when it shows declines, though the size of these changes varies across subgroups. In contrast, we observe mixed subgroup trends and lower $\dlift$ in MC-MED. }
    \label{fig:subgroup_lift}
\end{figure}

In \Cref{fig:subgroup_lift}, we visualize $\dlift$ across subgroups in both datasets.
Although the subgroup performance gains vary, the subgroup directional trend generally follows the overall task-level waveform \dlift from MIMIC-IV (see \cref{tab:agg-perf-and-lift}), with exceptions, such as  MedGemma-4B, which show both positive and negative \dlift\ across subgroups. 
 QoQ-Med shows a positive impact of ECG on \texttt{48-IHM} broadly across comorbidity, age, sex, race, admission type, and insurance groups, suggesting that the aggregate benefit is not driven by a narrow subgroup.
Similarly, MedGemma-27B shows a positive pattern for \texttt{48-IHM} across subgroups with significance.
However, on MC-MED, where aggregate $\dlift$ is small, subgroup trends are more mixed, with gains in some groups and declines in others.

\begin{wraptable}{r}{0.45\columnwidth}
\vspace{-13pt}
\centering
\scriptsize
\setlength{\tabcolsep}{1pt}
\setlength{\aboverulesep}{1pt}
\setlength{\belowrulesep}{1pt}
\renewcommand{\arraystretch}{1.5}
\captionsetup{font=small,skip=2pt}

\resizebox{\linewidth}{!}{%
\begin{tabular}{@{}lcr@{}}
\toprule
Model / task & Subgroup & $\dlift$ \\
\midrule
\multirow{2}{*}{\shortstack[l]{\texttt{QoQ-Med} /\\\texttt{48-IHM}}}
& Age & $+4.24^{*} \to +4.84^{*} \to +5.96^{*} \to +7.75^{*}$ \\
& CCI & $+3.04^{*} \to +3.97^{*} \to +7.68^{*} \to +10.07^{*}$ \\

\hdashline[1pt/1pt]
\texttt{QoQ-Med} /\texttt{LOS}
& Age & $-1.46 \to +1.39 \to -1.11 \to +3.29^{*}$ \\
\midrule
\renewcommand{\arraystretch}{0.95}

Subgroup & & Mean $\rho$  \\
\midrule
\texttt{CCI}   & & $+0.18$ \\
\texttt{Age}   &  & $+0.27$  \\
\bottomrule
\end{tabular}%
}

\caption{Macro-F1 lift across increasing age or CCI bins.
Mean Spearman $\rho$ measures association.}
\label{tab:trend-snapshot}
\vspace{-25pt}
\end{wraptable}

\textbf{Trends.} We analyze monotonic trends in $\dlift$ across subgroups along ordinal axes.
On MIMIC-IV \texttt{48-IHM}, QoQ-Med's multimodal lift is positive across every comorbidity bin and age and increases monotonically 
from $+4.24$\% for patients aged $<40$ to $+7.75$\% for patients aged $75+$ and approximately $+10$\% increase in the highest-CCI cohort.
Additionally, we observe a ``cross-over" multimodal performance pattern: using waveforms in addition to health records is unhelpful in lower-risk strata but helpful in higher-risk ones.
For instance, on MIMIC-IV \texttt{LOS}, adding waveforms to MedGemma-4B lowers macro-F1 in the younger cohort, but yields gains  with increasing age reaching $+3.29$. 

Averaged across all model\(\times\)task combinations, \(\dlift\) shows some positive associations with comorbidity burden (CCI) and age (see \cref{tab:trend-snapshot}).
These trends are suggestive that EHR+ECG multimodal improvements may be more useful in clinically relevant subgroups in these model--task settings; however, this pattern is not universal across models and tasks.

\begin{takeaway}{Takeaway}
Overall, \textbf{RQ2} shows that subgroup modality lift generally follows the aggregate direction on MIMIC-IV, for both gains and declines, although its magnitude varies across groups. MC-MED shows lower aggregate conditional modality lift and more mixed subgroup trends. This suggests that future work could benefit from subgroup-aware training methods to improve the consistency of multimodal gains across patient groups and clinical settings. 
\end{takeaway}

\subsection{RQ3: Cross-Modal Attention and Internal Representation Analysis}
\label{sec:rq3}
In this section, we assess how MedLLMs allocate attention between EHR text and ECG inputs using the content-normalized ECG attention share discussed in \Cref{sec:analysis_metrics}.
We examine how this allocation varies across model layers and use linear probes on layer-wise representations to assess how much task-relevant information can be decoded from each modality's representations.

\newcommand{\attnCIinline}[3]{%
  #1~{\tiny\textcolor{black!50}{[#2, #3]}}%
}

\begin{figure*}[t]
\centering

\begin{minipage}[c]{0.54\textwidth}
\centering
\footnotesize
\setlength{\tabcolsep}{2pt}
\renewcommand{\arraystretch}{1.6}
\captionsetup{skip=4pt}
\begin{tabular*}{\linewidth}{@{\extracolsep{\fill}}llccc@{}}
\toprule
\textbf{Model} & \textbf{Task} &
\textbf{ECG Helped} & \textbf{ECG Hurt} & $\Delta_{He-Hu}$ \\
\midrule
\multirow{2}{*}{QoQ-Med}
& \texttt{LOS}
& \attnCIinline{42.7}{42.5}{43.0}
& \attnCIinline{43.8}{43.4}{44.2}
& $-1.0$ \\
& \texttt{48-IHM}
& \attnCIinline{38.1}{38.0}{38.2}
& \attnCIinline{37.6}{37.4}{37.8}
& $+0.5$ \\
\addlinespace[3pt]
\multirow{2}{*}{MedGemma-4B}
& \texttt{LOS}
& \attnCIinline{32.4}{31.9}{33.0}
& \attnCIinline{35.2}{34.7}{35.6}
& $-2.8$ \\
& \texttt{48-IHM}
& \attnCIinline{29.6}{29.1}{30.1}
& \attnCIinline{33.6}{33.4}{33.9}
& $-4.0$ \\
\midrule
\textbf{Average} & All
& \attnCIinline{35.7}{35.4}{36.1}
& \attnCIinline{37.6}{37.2}{37.9}
& $-1.8$ \\
\bottomrule
\end{tabular*}

\captionof{table}{Mean ECG attention share (\%) on MIMIC-IV
[95\% bootstrap CI]. $\Delta_{He-Hu}$ is helped minus hurt. }
\label{tab:ecg-attention-helps}
\end{minipage}
\hfill
\begin{minipage}[c]{0.44\textwidth}
\centering
\captionsetup{skip=4pt}

\includegraphics[width=\linewidth]
  {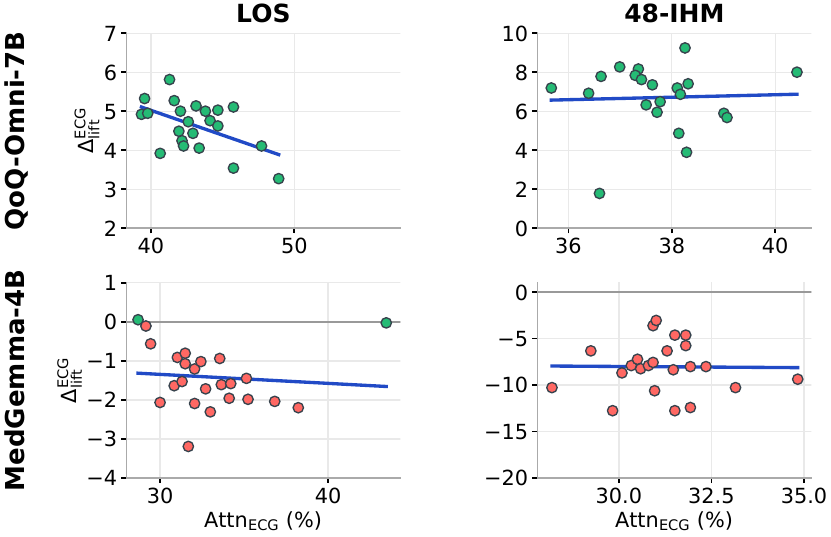}

\captionof{figure}{Subgroup-average ECG attention share
versus subgroup-specific $\dlift$. }
\label{fig:attn-vs-subgroup-lift}
\end{minipage}

\end{figure*}
\textbf{Attention Share Analysis.}
We observe that for the MIMIC-IV tasks, ECG attention share is non-trivial across both the evaluated models (see average  in \cref{tab:ecg-attention-helps})).
In \Cref{fig:layerwise-attn} (Appendix) we observe that earlier full-layer attention favors ECG, whereas later layers favor EHR text on \texttt{48-IHM}.
However, the aggregate attention share across all layers on signal tokens remains below $50\%$, indicating a larger average attention share for EHR text, consistent with text modality dominance patterns reported in prior VLM  studies \cite{parcalabescu2025do,zhang2025evaluating,restrepo2025risk}.

\textbf{Outcome and Subgroup Comparisons.}
We compare attention to ECG across cases where its inclusion corrects an erroneous EHR-only prediction (i.e, \textit{ECG Helps}) and cases where it leads to an incorrect prediction despite a correct EHR-only prediction (i.e, \textit{ECG Hurts}).
From \cref{tab:ecg-attention-helps}, we observe that ECG attention is not consistently higher when ECG helps, and the direction of the difference varies across evaluated model--task combinations. 
Across patient subgroups, higher ECG attention also does not consistently correspond to greater $\dlift$ in \cref{fig:attn-vs-subgroup-lift}.
For example, the fitted trend is negative for QoQ-Med-Omni-7B on \texttt{LOS} but weakly positive on \texttt{48-IHM}. MedGemma shows the opposite pattern across these tasks.

\begin{wrapfigure}{r}{0.45\columnwidth}
\vspace{-12pt}
\centering
\begin{minipage}{\linewidth}
\centering
\footnotesize
\renewcommand{\arraystretch}{1.1}
\begin{tabular*}{\linewidth}{@{\extracolsep{\fill}}lccc@{}}
\toprule
Layer & EHR & ECG & Fused \\
\midrule
0 & 0.793 & 0.617 & 0.758 \\
19 & 0.827 & 0.560 & 0.789 \\
33 & 0.794 & 0.555 & 0.789 \\
\bottomrule
\end{tabular*}%
\captionof{table}{Linear-probe AUROC at selected layers of MedGemma-4B on \texttt{48-IHM}.}
\label{tab:probe-main}
\end{minipage}
\vspace{-10pt}
\end{wrapfigure}

The attention results show that higher ECG attention does not consistently correspond to improved predictions, consistent with prior work cautioning against interpreting attention as feature importance or prediction faithfulness \citep{jain2019attention,abnar2020quantifying}  and with broader LVLM studies on modality imbalance, where non-text inputs may be underused, overemphasized, or inefficiently converted into useful predictions \citep{parcalabescu2025do,zhang2025evaluating,wu2025languageoverrules}.

\textbf{Layer-wise Linear Probing.}
We analyze cross-modal layer-wise representations in MedGemma-4B. At selected intermediate decoder layers, we mean-pool representations extracted from the residual stream \cite{elhage2021mathematical} separately over the ECG-image and EHR-text token spans. We use these hidden representations to train linear probes for the \texttt{48-IHM} task.
We observe in \cref{tab:probe-main} that ECG-probe AUROC declines overall from 0.617  at the input layer to 0.555 at layer 33, whereas EHR representations remain strongly discriminative, reaching 0.827 at layer 19.
Fused multimodal feature-probe AUROC increases overall with depth but does not outperform EHR-only probing. 
These results suggest that while the initial ECG representations encode some task-relevant information, the linear decodability of this information decreases across the model's layers. Furthermore, for the evaluated task, combining the extracted ECG and EHR representations does not improve outcome discrimination. Thus, layer-wise probing reveals differences in the evolution of task-relevant information across modalities with depth.

\begin{takeaway}{Takeaway}
Overall, \textbf{RQ3} helps us understand cross-modal processing in MedLLMs, where we observe text dominance in attention allocation, while ECG attention share does not consistently track predictive benefit.
Linear probing further shows that combining ECG- and EHR-token representations does not improve outcome discrimination over EHR-token representations alone.
\end{takeaway}

\begin{wrapfigure}{r}{0.35\columnwidth}
 \vspace{-10pt}
    \centering
    \includegraphics[width=\linewidth]{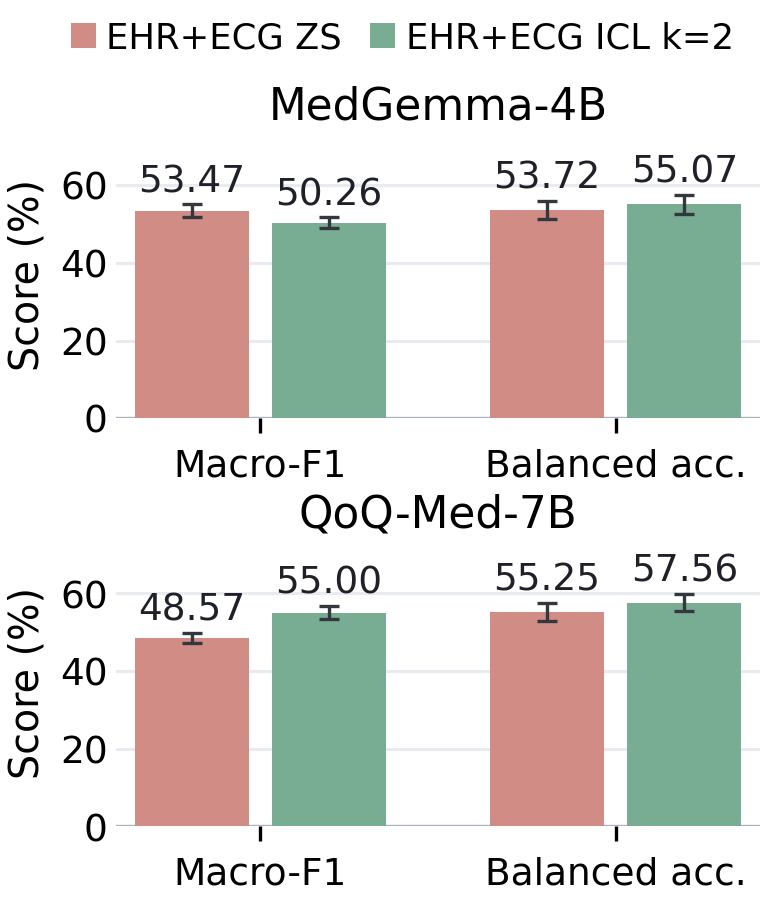}
    \caption{Multimodal in-context learning performance on
    \texttt{48-IHM}.
    }
    \label{fig:icl-two-shot}
    \vspace{-10pt}
\end{wrapfigure}

\subsection{RQ4: Multimodal In-Context Learning}
\label{sec:icl}
While task-specific fine-tuning and other post-training approaches offer avenues for adaptation to new clinical settings, our zero-shot evaluation focuses on current MedLLMs’ out-of-the-box capabilities. A natural next question is whether few-shot in-context learning (ICL), which provides labeled examples in the prompt without updating model parameters, can improve multimodal (EHR+ECG) performance.
We explore in-context learning (ICL) by prepending two labeled EHR+ECG demonstrations to the frozen model on \texttt{48-IHM}. 
On this imbalanced task, QoQ-Med's macro-F1 increases from $48.57$ to $55.00$ (see Figure~\ref{fig:icl-two-shot}), accompanied by increased majority-class recall. 
However, for MedGemma-4B, macro-F1 decreases by $\approx3\%$.  Balanced accuracy increases by $+2\%$ points for both models. 
For MedGemma-4B, higher balanced accuracy and lower macro-F1 accompany improved rare-class recall and reduced majority-class recall.
These responses show that a small demonstration context can help with multimodal performance. 
These different MedLLM family behaviors in our findings motivate further investigation of demonstration-based adaptation for multimodal clinical prediction with MedLLMs. 

\begin{takeaway}{Takeaway}
Overall, \textbf{RQ4} shows that the effect of multimodal in-context demonstrations can benefit some tasks but varies across LLM families, highlighting the potential for lightweight, model-specific adaptation using demonstrations.
\end{takeaway}

\section{Conclusion}
In this paper, we presented \wav, a systematic, zero-shot evaluation of how well open-source medical LLMs jointly model structured EHR data and physiological waveforms, including 12-lead ECG and bedside ECG+PPG, for clinical outcome prediction.
We measured the conditional modality lift $\dlift$, i.e., the difference in performance obtained by using both EHR-waveform inputs instead of EHR-only inputs, and complemented this with subgroup-stratified and attention-based analyses, counterfactual waveform controls, and linear probing.
We find that the predictive benefit of adding waveforms in the zero-shot setting varies across models, tasks, and patient subgroups.
Overall, \wav highlights the multimodal capabilities of MedLLMs and opportunities to improve how they integrate clinically relevant, widely available waveform data with structured EHRs.

\newpage
\bibliographystyle{plain}
\bibliography{ref}
\appendix

\clearpage
\appendix
\section*{\Large Appendix}
\section{Dataset Sources and Access}
\label{app:data}

\subsection{MIMIC-IV}
We use the MIMIC-IV dataset (Johnson et al., 2023)~\citep{mimiciv} together with MIMIC-IV-ECG (Gow et al., 2023)~\citep{gow2023mimic} which provides 10-second 12-lead diagnostic electrocardiograms linked to MIMIC-IV admissions. Both resources are distributed by PhysioNet \cite{goldberger2000physiobank} under the \emph{PhysioNet Credentialed Health Data License 1.5.0}, which prohibits sharing access with parties not under the same Data Use Agreement and prohibits any attempt to re-identify individuals or institutions.
We do not redistribute MIMIC-IV or MIMIC-IV-ECG, in whole or in any subset, or in any derivative form to any third party.  Credentialed users can access these datasets on PhysioNet.

\subsection{MC-MED}
We use the Multimodal Clinical Monitoring in the Emergency Department (MC-MED) dataset~\citep{mcmed}, which provides emergency-department encounters paired with continuous bedside-monitor physiological waveforms (single-lead ECG (\texttt{II}), photoplethysmogram (\texttt{Pleth}), and respiration (\texttt{Resp})), structured EHR (laboratory results, orders, numerics, radiology reports, past medical history, medications, and discharge diagnoses), and demographic attributes. MC-MED is distributed by PhysioNet under the same Credentialed Health Data License and is subject to the same restrictions described above.

\section{Tasks}
\label{app:tasks}

We define each task by an \emph{observation window} for model inputs and a
\emph{prediction window} for outcome ascertainment. Where the ECG lookback differs from the EHR observation window, we specify it separately.

\subsection{MIMIC-IV tasks}
We paired structured EHR data with diagnostic ECGs across two MIMIC-IV tasks, building on prior multimodal evaluations~\citep{dai2025climb,qoqmed}. 

\paragraph{\texttt{48-IHM}.}
We predicted survival to hospital discharge among admissions with an eligible ECG, and at least 48 hours of inpatient observation. 
The model received demographics, admission information, laboratory results, charted measurements, and medications from the first 48 hours. We paired these data with the ECG closest to admission among recordings acquired from
24 hours before admission through hour 48. 
The label was \texttt{no} if an in-hospital death was recorded after hour 48 and \texttt{yes} otherwise.

\paragraph{\texttt{LOS}.}
We predicted total hospital length of stay from structured EHR data and one diagnostic ECG acquired during the first 48 hours after admission.
Eligible admissions had at least 48 hours of observation and a recorded discharge time. 
Following QoQ-Med~\citep{qoqmed}, we assigned four classes based on total length of stay, with \texttt{A} for up to 4 days, \texttt{B} for 5--8 days, \texttt{C}
for 9--12 days, and \texttt{D} for more than 12 days.

\subsection{MC-MED tasks}
We adapted two tasks from the
Multimodal Clinical Benchmark for Emergency Care
(MC-BEC)~\citep{chen2023multimodal}. 
These tasks let us assess whether bedside
waveforms complement the EHR at different time horizons, from short-term physiologic deterioration to  return visits. 
For ED-Revisit, we required at least one waveform file for
each of ECG lead \texttt{II}, PPG \texttt{Pleth}, and respiration \texttt{Resp}.
For Decompensation, waveform eligibility was determined at the prediction time.

\paragraph{\texttt{ED-Decompensation}.}
We predicted short-term physiologic deterioration one hour after rooming, using ED arrival as the time anchor when rooming time was unavailable. 
The model received structured EHR data through the prediction time and ECG or PPG data from the observation window. Post-prediction vital signs and discharge diagnoses were excluded from the model inputs.
Eligible visits had baseline heart rate and oxygen saturation measurements, no deterioration criterion met during the preceding hour or at the cutoff, and a waveform covering the required time interval. 
Following the criteria in \cite{chen2023multimodal}, the label was \texttt{yes} if heart rate
exceeded 110 bpm, mean arterial pressure fell below 65 mmHg, or peripheral oxygen saturation fell below 90\% during the two hours after prediction, with follow-up ending earlier if monitoring stopped. The label was \texttt{no} otherwise.

\paragraph{\texttt{ED-Revisit}.}
We used structured EHR data and waveforms from the index visit
through ED departure to predict the time to the next recorded ED visit.
Using the 3-, 7-, and 14-day horizons from MC-BEC~\citep{chen2023multimodal},
we assigned four mutually exclusive classes. 
We assigned \texttt{bin\_0\_3}
for a return within 3 days, \texttt{bin\_3\_7} for a return after 3 and
within 7 days, and \texttt{bin\_7\_14} for a return after 7 and within
14 days. Visits with a return after 14 days or no recorded return were
assigned \texttt{bin\_none} (majority class).

\section{Data Preparation Pipeline}
\label{sec:ehr_text_construction}
\subsubsection{MIMIC-IV}

For each admission, we create a single patient narrative that begins with demographic information (age and sex), followed by the current admission metadata, the procedures performed during the admission, a chronologically ordered stream of laboratory results, charted vital signs and bedside measurements, and the patient's active medication list.
For models with a native time-series interface (QoQ-Med), we provide the 10-second 12-lead diagnostic ECG as a 8-lead tensor of shape $[B, 8, 2500]$ following the QoQ-Med encoder's expected input format.
For models that ingest physiological signals as images, the same 8-lead trace is rendered as a $4 \times 2$ matplotlib panel.

\subsubsection{MC-MED}

For each ED encounter, we render a structured narrative that begins with a visit-summary header (demographics, arrival information, and triage findings), followed by the patient's prior medical history, the medications they were taking at arrival, the diagnoses associated with the current visit, and a chronologically ordered event log of laboratory results, clinical orders, bedside numeric measurements, and radiology study metadata.
Information that is only available after the prediction point is excluded so the narrative reflects only what was available at prediction time.
We extract 10-second single-lead ECG and single-channel PPG segments from the WFDB waveform records from the observation window.
Although longer recordings can provide additional temporal context \cite{gustafsson2026signalmc}, we limit the input duration to accommodate LLM's vision/time-series modality encoder constraints.
We use the same waveform duration across MedLLMs and supervised fusion baselines to ensure a consistent comparison.

\subsubsection{Subgroups}
\label{sec:subgroups}
Subgroups are defined along axes drawn from the dataset metadata, following
prior work on multimodal clinical evaluation~\cite{meng2022interpretability,chen2023multimodal}:
demographic (age band, sex, race, insurance), clinical (Charlson
Comorbidity Index~\cite{charlson1987new}, admission type), and temporal
(anchor-year window) for MIMIC-IV; for MC-MED we use the same
demographic and CCI axes, plus the ED-specific means of arrival and triage acuity (Emergency Severity Index).

\begin{table*}[h]
\centering
\footnotesize
\setlength{\tabcolsep}{5pt}
\renewcommand{\arraystretch}{1.2}
\begin{tabular}{@{}lllcrl@{}}
\toprule
\textbf{Task} & \textbf{Dataset} & \textbf{Signal} & \textbf{Cls.} & \textbf{$n$} & \textbf{Label counts} \\
\midrule
48-IHM & MIMIC-IV & ECG & 2 & 5,420 & Survived: 5,000; died: 420 \\
\addlinespace[4pt]
LOS$^{\ddagger}$ & MIMIC-IV & ECG & 4 & 4,573 &
{\setlength{\tabcolsep}{4pt}
\begin{tabular}[t]{@{}cccc@{}}
$<5$ d & 5--8 d & 9--12 d & $>12$ d \\
\cmidrule(lr){1-4}
2,079 & 1,395 & 535 & 564
\end{tabular}} \\
\midrule
ED-Decomp & MC-MED & ECG & 2 & 2,338 & No: 1,964; yes: 374 \\
          &        & PPG & 2 & 2,831 & No: 2,425; yes: 406 \\
\addlinespace[4pt]
ED-Revisit$^{\ddagger}$ & MC-MED & ECG & 4 & 3,084 &
{\setlength{\tabcolsep}{4pt}
\begin{tabular}[t]{@{}cccc@{}}
None & 0--3 d & 3--7 d & 7--14 d \\
\cmidrule(lr){1-4}
2,723 & 157 & 101 & 103
\end{tabular}} \\
\addlinespace[3pt]
                      &        & PPG & 4 & 3,083 &
{\setlength{\tabcolsep}{4pt}
\begin{tabular}[t]{@{}cccc@{}}
None & 0--3 d & 3--7 d & 7--14 d \\
\cmidrule(lr){1-4}
2,722 & 157 & 101 & 103
\end{tabular}} \\
\bottomrule
\end{tabular}
\caption{Cohort sizes and class distributions; $^{\ddagger}$cardiac subgroup.}
\label{tab:tasks}
\end{table*}

\begin{table}[h]
\centering
\small
\caption{MLLMs evaluated; physiological-signal ingestion mechanism; and attention regimes}
\label{tab:models}
\begin{tabularx}{\textwidth}{@{} p{3.0cm} p{1.5cm}Y Y@{}}
\toprule
\textbf{Model} & \textbf{Params} & \textbf{Signal ingestion} & \textbf{Attention} \\
\midrule
QoQ-Med-7B \cite{qoqmed}
& 7B-class
& Native 8-lead ECG time-series encoder, $[B,8,2500]$
& Full causal \\

\addlinespace
MedGemma-4B\cite{sellergren2025medgemma}
& 4B
& Rendered as a $4 \times 2$ ECG image
& 5 sliding window + 1 full \\

\addlinespace
MedGemma-27B \cite{sellergren2025medgemma}
& 27B
&  Rendered as a $4 \times 2$ ECG image
&  5 sliding window + 1 full \\

\bottomrule
\end{tabularx}
\end{table}
\section{Models}
\label{app:models}

We evaluate two open-weight medical multimodal LLMs that differ in (i) the language backbone, (ii) the visual/physiological encoder, and (iii) how a physiological signal enters the model.

\subsection{QoQ-Med}
QoQ-Med-7B \citep{qoqmed} is a 7B-parameter clinical multimodal model (checkpoint used \href{https://huggingface.co/ddvd233/QoQ-Med-Omni-7B}{\texttt{ddvd233/QoQ-Med-Omni-7B}}). It is built on a Qwen2.5-family backbone, post-trained for clinical prediction, and trained with domain-aware GRPO on a mixture of medical text, medical imaging (CXR, dermatology, ophthalmology), and 12-lead ECG time series. It incorporates a \emph{native time-series interface}: ECG enters through a dedicated time-series encoder which is trained to be aligned with LLM.

\subsection{MedGemma}
MedGemma-4B-it \citep{sellergren2025medgemma} is a 4B-parameter multimodal medical LLM (checkpoint used \href{https://huggingface.co/google/medgemma-4b-it}{\texttt{google/medgemma-4b-it}}.) It is built on the Gemma 3 multimodal architecture (decoder LLM + SigLIP-style vision encoder) and continued-pretrained / instruction-tuned on medical text and medical images (radiology, pathology, dermatology, fundus). MedGemma does not have a native physiological-signal interface; we therefore route ECG through the standard image path, which is then ingested by the SigLIP vision encoder. The Gemma 3 decoder uses interleaved sliding-window and full-attention layers; for the attention-share metric, we restrict averaging to full-attention layers, where EHR and signal spans are jointly visible.

All models are run in inference-only mode, with greedy decoding, bfloat16 weights, and FlashAttention-2. Maximum new tokens are capped at 512 for MedGemma across all tasks; for QoQ-Med we use the upstream-recommended 4{,}096 tokens.

\section{Representative Prompt Templates}
\label{app:prompt}

We show the 48-IHM prompts for both models as representative examples.

\begin{prompttemplatebox}{MIMIC-IV: 48-IHM --- QoQ-Med}
Below is a history of a patient: \\
\{ehr\_text\} \\
\\
Based on the information above from the first 48 hours of admission,
will this patient survive this hospital admission?
Answer with yes or no. \\
You FIRST think about the reasoning process as an internal monologue
and then provide the final answer. The reasoning process MUST BE
enclosed within \texttt{<think>} \texttt{</think>} tags.
The final answer MUST BE put in
\texttt{\textbackslash boxed\{\}}.
\end{prompttemplatebox}

\begin{prompttemplatebox}{MIMIC-IV: 48-IHM --- MedGemma}
\textbf{[SYSTEM]} \\
You are a helpful medical AI assistant. Please be concise. \\
\\
\textbf{[USER]} \\
Below is a history of a patient: \\
\{ehr\_text\} \\
\\
Will this patient survive this hospital admission?
Answer with yes or no. \\
\\
IMPORTANT: End your response with exactly: Final Answer: (X)
\end{prompttemplatebox}

We summarize the task-specific instructions  for the oether tasks below.
Each prompt includes the patient's EHR context.
Repeated system, reasoning, and response-format instructions
are omitted from the boxes for brevity and same as above.

\begin{prompttemplatebox}{ LOS}
How long will the patient stay in the hospital?
Choose exactly one: \\
A. 0--4 days \\
B. 5--8 days \\
C. 9--12 days \\
D. more than 12 days
\end{prompttemplatebox}

\vspace{10pt}
\begin{prompttemplatebox}{ED-Decomp}
The patient did not cross a decompensation threshold during
the first hour of monitoring. Based on the ED data available
so far, predict whether the patient will newly decompensate
within the next 120 minutes. \\
\\
Decompensation is the new onset of any of:
HR $>110$, MAP $<65$ mmHg, or SpO$_2<90\%$. \\
Choose exactly one:
\texttt{no\_decomp} or \texttt{decomp}.
\end{prompttemplatebox}

\vspace{10pt}
\begin{prompttemplatebox}{ED-Revisit}
Predict the most likely ED revisit outcome after discharge.
Choose exactly one: \\
\texttt{bin\_0\_3}: revisit within 0--3 days \\
\texttt{bin\_3\_7}: revisit within 3--7 days \\
\texttt{bin\_7\_14}: revisit within 7--14 days \\
\texttt{bin\_none}: no revisit within 14 days 
\end{prompttemplatebox}
\vspace{10pt}

\section{Supervised baselines}
\label{app:supervised_baselines}
\begin{table*}[t]
\centering
\small
\caption{Supervised baselines using frozen encoders. All methods are evaluated on the same test set of $n$ waveform-present encounters within each task. The majority baseline always predicts the most frequent class. AUROC is positive-class for binary tasks and macro one-vs-rest for multiclass tasks. Bold indicates the highest score among the four learned baselines for each task and metric.}
\label{tab:sup-all}
\setlength{\tabcolsep}{5pt}
\renewcommand{\arraystretch}{1.5}
\resizebox{\textwidth}{!}{%
\begin{tabular}{@{}ll *{6}{c}@{}}
\toprule
& & \multicolumn{2}{c}{\textbf{MIMIC-IV}}
    & \multicolumn{4}{c}{\textbf{MC-MED}} \\
\cmidrule(lr){3-4}
\cmidrule(lr){5-8}
\textbf{Metric} & \textbf{Method}
& \shortstack{\texttt{48-IHM} \\ \\ECG}
& \shortstack{\texttt{48h-LOS}\\ \\ECG}
& \shortstack{\texttt{ED-Decomp}\\ \\ECG}
& \shortstack{\texttt{ED-Decomp}\\  \\PPG}
& \shortstack{\texttt{ED-Revisit}\\  \\ECG}
& \shortstack{\texttt{ED-Revisit}\\  \\PPG} \\
\midrule

\multirow{5}{*}{AUROC}
& Majority
& 0.500 & 0.500 & 0.500 & 0.500 & 0.500 & 0.500 \\
& EHR only
& 0.714 & 0.570 & 0.579 & 0.583 & 0.583 & \textbf{0.592} \\
& Waveform only
& 0.712 & 0.558 & 0.619 & 0.619 & 0.506 & 0.517 \\
\arrayrulecolor{gray!55}
\cdashline{2-8}
\arrayrulecolor{black}
& Late fusion
& \textbf{0.764} & \textbf{0.587} & \textbf{0.633}
& \textbf{0.641} & 0.586 & \textbf{0.592} \\
& Concatenation
& 0.736 & 0.575 & 0.627 & 0.632 & \textbf{0.593} & 0.586 \\

\arrayrulecolor{gray!65}
\midrule
\arrayrulecolor{black}

\multirow{5}{*}{Macro-F1}
& Majority
& 0.480 & 0.156 & 0.457 & 0.461 & 0.235 & 0.235 \\
& EHR only
& 0.531 & 0.272 & 0.501 & 0.498 & 0.238 & 0.211 \\
& Waveform only
& 0.528 & 0.256 & 0.521 & 0.516 & 0.178 & 0.223 \\
\arrayrulecolor{gray!55}
\cdashline{2-8}
\arrayrulecolor{black}
& Late fusion
& 0.559 & 0.278 & 0.530 & \textbf{0.531} & 0.238 & 0.220 \\
& Concatenation
& \textbf{0.591} & \textbf{0.284} & \textbf{0.551}
& 0.526 & \textbf{0.290} & \textbf{0.241} \\

\arrayrulecolor{gray!65}
\midrule
\arrayrulecolor{black}

\multirow{5}{*}{Accuracy}
& Majority
& 0.923 & 0.455 & 0.840 & 0.857 & 0.883 & 0.883 \\
& EHR only
& 0.712 & 0.302 & 0.591 & 0.606 & 0.528 & 0.402 \\
& Waveform only
& 0.713 & 0.272 & 0.615 & 0.619 & 0.348 & 0.502 \\
\arrayrulecolor{gray!55}
\cdashline{2-8}
\arrayrulecolor{black}
& Late fusion
& 0.751 & 0.303 & 0.623 & 0.636 & 0.529 & 0.440 \\
& Concatenation
& \textbf{0.832} & \textbf{0.312} & \textbf{0.675}
& \textbf{0.641} & \textbf{0.789} & \textbf{0.533} \\

\bottomrule
\end{tabular}%
}
\arrayrulecolor{black}
\end{table*}

We evaluate supervised baselines to assess the predictive value of EHR and waveform features, both individually and jointly.
EHR features are extracted using a frozen ClinicalBERT encoder~\cite{huang2019clinicalbert}, taking the final-layer \texttt{[CLS]} embedding of text constructed from the same structured EHR variables supplied to the MedLLMs.
For waveform features, we use frozen ECG-JEPA~\cite{kim2024learning} and PaPaGei-P~\cite{pillai2024papagei} encoders for ECG and PPG, respectively, mean-pooling the ECG-JEPA patch representations into a single embedding.
For each task--waveform setting, we train class-weighted linear probes on the EHR and waveform embeddings separately.
We then evaluate two fusion strategies: \textit{late fusion}, which blends the class probabilities from the two probes, and \textit{concatenation}, which trains an MLP on the concatenated embeddings.
All prediction heads are trained on the training split, with head hyperparameters and the late-fusion weight selected on a held-out validation set.

\paragraph{Multimodal Supervised Gains.}
We compare each fusion strategy with the corresponding EHR-only probe to assess the additional predictive value of waveforms.
As shown in \Cref{tab:sup-all}, concatenation improves macro-F1 across all six task--waveform settings.
These gains span across tasks and include both ECG and PPG, suggesting that physiological signals can provide useful information under task-specific supervision.

\textbf{MIMIC-IV:}
On \texttt{48-IHM}, multimodal concatenation increases macro-F1 from $53.1\%$ to $59.1\%$ and achieves the highest accuracy, while late fusion achieves the highest AUROC, improving from $0.714$ to $0.764$.
On \texttt{LOS}, the gains are smaller, but fusion still improves macro-F1 by $1.2$ percentage points and increases AUROC over the EHR-only baseline.
The waveform-only ECG probes also achieve AUROCs above chance ($0.5$), reaching $0.712$ on \texttt{48-IHM} and $0.558$ on \texttt{LOS}, suggesting that ECG features can capture task-relevant discriminative information in the evaluated cohorts.

\textbf{MC-MED:}
On \texttt{ED-Decomp}, ECG concatenation improves macro-F1 by $5.0$ percentage points, while PPG late fusion improves it by $3.3$ points.
The waveform-only probes achieve an AUROC of $0.619$ for both ECG and PPG, exceeding the corresponding EHR-only probes ($0.579$ and $0.583$, respectively), suggesting that either signal alone supports discrimination in the evaluated cohorts.
Late fusion further increases AUROC for both signals, reaching $0.633$ with ECG and $0.641$ with PPG.
Together with the results on \texttt{48-IHM}, these improvements demonstrate benefits in both classification and ranking performance under supervised fusion.
On \texttt{ED-Revisit}, concatenation improves macro-F1 by $5.2$ points with ECG and $3.0$ points with PPG, largely through improved majority-class performance.
Although waveform-only AUROCs are close to chance, ECG concatenation increases AUROC over EHR alone.
For PPG, fusion does not improve AUROC over EHR alone, indicating that the observed benefit is specific to classification performance.

These task-specific supervised baselines, evaluated on the same test split as the MedLLMs, provide a reference for interpreting zero-shot MedLLM performance. \textit{The EHR-only baselines help assess how effectively MedLLMs use the predictive information in structured EHR data, while the fusion baselines allow us to examine whether MedLLMs obtain comparable incremental benefits from adding waveforms.}

\begin{figure}[h]
    \centering
    \includegraphics[width=1.0\linewidth]{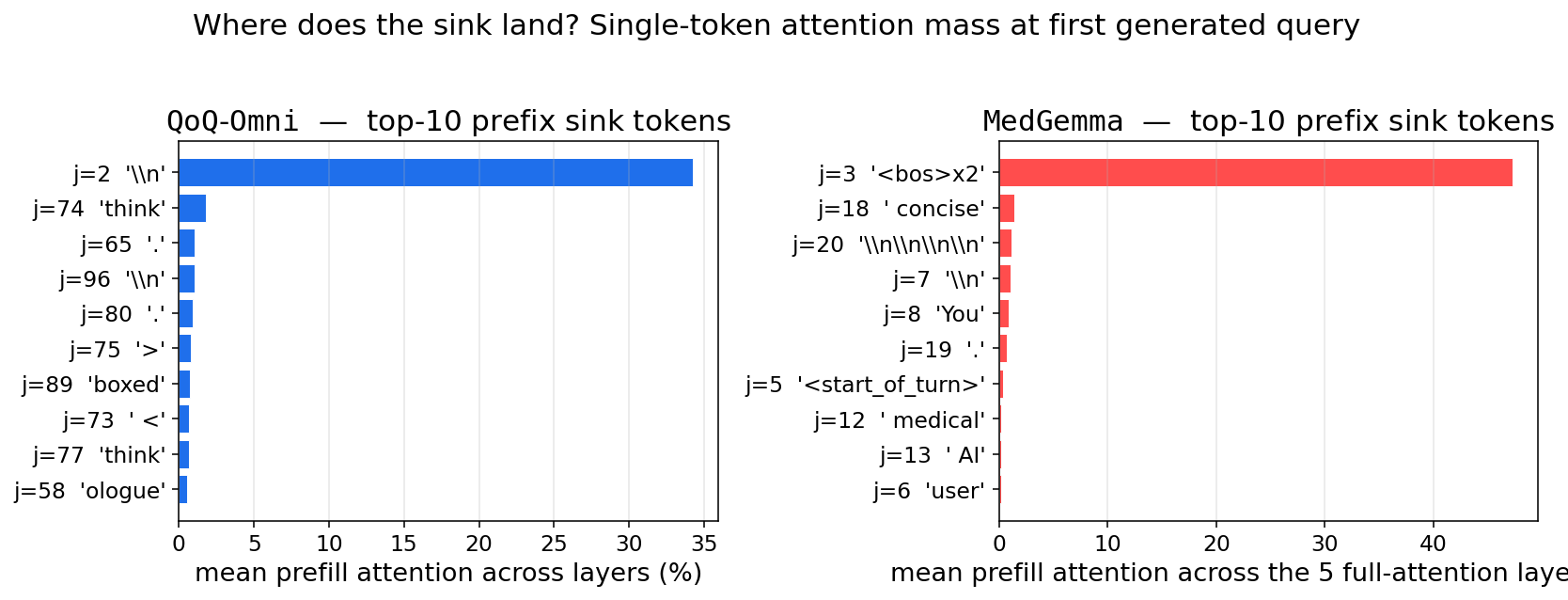}
    \caption{Attention Sink Behavior in Qwen based QoQ and MedGemma MedLLMs}
    \label{fig:attention-sink}
\end{figure}
\section{Layer-Wise Cross-Modal Attention and Representations Linear Probing}
We present additional analyses of layer-wise attention, linear probing of modality-specific representations, and attention-share comparisons across prediction outcomes and patient subgroups.
We examine attention allocation and modality-specific representations across layers of MedGemma-4B on \texttt{48-IHM}.
The attention analysis uses a balanced subset of 200 patients, while the probing analysis uses the full test set as in \cref{tab:data}.

\begin{figure*}[h]
\centering

\begin{minipage}[t]{0.48\textwidth}
\vspace{0pt}
\centering
\footnotesize
\setlength{\tabcolsep}{3pt}
\renewcommand{\arraystretch}{1.3}
\captionsetup{font=footnotesize,skip=3pt}

\begin{tabular*}{\linewidth}{@{\extracolsep{\fill}}lccc@{}}
\toprule
Layer & EHR & ECG & Fused \\
\midrule
0  & 0.793 & 0.617 & 0.758 \\
9  & 0.795 & 0.605 & 0.765 \\
19 & 0.827 & 0.560 & 0.789 \\
25 & 0.809 & 0.573 & 0.795 \\
33 & 0.794 & 0.555 & 0.789 \\
\bottomrule
\end{tabular*}%

\captionof{table}{Linear-probe AUROC at representative
layers of MedGemma-4B on \texttt{48-IHM}.
Fused denotes concatenated EHR and ECG representations.}
\label{tab:probe}
\end{minipage}
\hfill
\begin{minipage}[t]{0.48\textwidth}
\vspace{0pt}
\centering
\captionsetup{font=footnotesize,skip=3pt}

\includegraphics[width=0.8\linewidth]{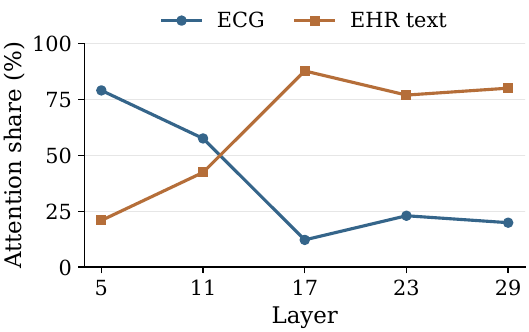}

\captionof{figure}{Two-way ECG and EHR attention shares
across full-attention layers of MedGemma-4B on \texttt{48-IHM}.}
\label{fig:layerwise-attn}
\end{minipage}

\end{figure*}

\textbf{Layer-wise attention.}
\Cref{fig:layerwise-attn} shows the two-way attention share allocated to ECG and EHR text across full-attention layers.
Earlier layers predominantly attend to ECG, whereas later layers favor EHR text.
Averaged across these layers, ECG receives less attention than EHR tokens.

\textbf{Layer-wise probing.}
We fit linear probes to the EHR token representations, ECG representations, and their concatenation at selected layers.
As shown in \cref{tab:probe}, ECG representations are most discriminative at the input layer (AUROC 0.617), with an overall decline across depth to 0.555 at layer 33.
EHR representations remain strongly discriminative, reaching an AUROC of 0.827 at layer 19.
Concatenating the representations does not outperform the EHR-only probe at any evaluated layer.
Fused-probe AUROC increases overall from 0.758 at the input layer to 0.795 at layer 25, before reaching 0.789 at layer 33.

\clearpage
\section{Additional Results}
In this section, we report additional aggregate metrics, subgroup-level performance differences, trend analysis  and  attention-performance analysis results blow.
Tables in this section report per-class conditional modality lift using various metrics such as specificity and sensitivity.

\begin{table}[h]
\centering
\scriptsize
\setlength{\tabcolsep}{4.2pt}
\renewcommand{\arraystretch}{1}

\label{tab:cond-lift-table}
\resizebox{\textwidth}{!}{%
\begin{tabular}{@{}llll ccc@{}}
\toprule
\textbf{Dataset} & \textbf{Task} & \textbf{Signal} & \textbf{Metric} & \textbf{QoQ-Med} & \textbf{MedGemma-4B} & \textbf{MedGemma-27B} \\
\midrule
\multirow{6}{*}{MIMIC-IV}
    & \multirow{3}{*}{\texttt{48-IHM}} & \multirow{3}{*}{ECG (12-lead)} & F1-Macro & \textbf{+6.12}\textsubscript{\textcolor{gray}{[+4.67,+7.46]}} & $-$0.38\textsubscript{\textcolor{gray}{[$-$2.48,+1.56]}} & \underline{+5.36}\textsubscript{\textcolor{gray}{[+4.18,+6.59]}} \\
    & & & Bal.\ acc. & +1.48\textsubscript{\textcolor{gray}{[$-$1.56,+4.75]}} & \underline{+3.27}\textsubscript{\textcolor{gray}{[+0.86,+5.41]}} & \textbf{+5.20}\textsubscript{\textcolor{gray}{[+2.73,+7.97]}} \\
    & & & Acc. & \textbf{+12.99}\textsubscript{\textcolor{gray}{[+11.22,+14.63]}} & $-$4.43\textsubscript{\textcolor{gray}{[$-$5.81,$-$2.86]}} & \underline{+9.59}\textsubscript{\textcolor{gray}{[+8.01,+11.01]}} \\
\cmidrule(l){2-7}
    & \multirow{3}{*}{\texttt{LOS}} & \multirow{3}{*}{ECG (12-lead)} & F1-Macro & \textbf{+1.24}\textsubscript{\textcolor{gray}{[$-$0.26,+2.95]}} & \underline{+0.10}\textsubscript{\textcolor{gray}{[$-$0.60,+1.08]}} & $-$2.56\textsubscript{\textcolor{gray}{[$-$4.06,$-$1.07]}} \\
    & & & Bal.\ acc. & \textbf{+0.70}\textsubscript{\textcolor{gray}{[$-$1.16,+2.71]}} & $-$1.65\textsubscript{\textcolor{gray}{[$-$2.36,$-$0.87]}} & \underline{$-$1.44}\textsubscript{\textcolor{gray}{[$-$2.92,$-$0.17]}} \\
    & & & Acc. & \textbf{+3.89}\textsubscript{\textcolor{gray}{[+2.32,+6.19]}} & \underline{$-$2.89}\textsubscript{\textcolor{gray}{[$-$3.72,$-$2.06]}} & $-$2.97\textsubscript{\textcolor{gray}{[$-$4.26,$-$1.71]}} \\
\midrule
\multirow{12}{*}{MCMED}
    & \multirow{3}{*}{\texttt{ED-Decomp}} & \multirow{3}{*}{ECG (bedside)} & F1-Macro & \textbf{$-$0.25}\textsubscript{\textcolor{gray}{[$-$0.50,+0.00]}} & \underline{$-$0.78}\textsubscript{\textcolor{gray}{[$-$2.26,+0.57]}} & $-$1.16\textsubscript{\textcolor{gray}{[$-$3.32,+1.11]}} \\
    & & & Bal.\ acc. & \textbf{+0.33}\textsubscript{\textcolor{gray}{[+0.15,+0.49]}} & \underline{$-$0.21}\textsubscript{\textcolor{gray}{[$-$0.99,+0.59]}} & $-$0.55\textsubscript{\textcolor{gray}{[$-$2.04,+1.05]}} \\
    & & & Acc. & \textbf{$-$0.79}\textsubscript{\textcolor{gray}{[$-$1.59,+0.04]}} & $-$1.39\textsubscript{\textcolor{gray}{[$-$2.28,$-$0.47]}} & \underline{$-$1.28}\textsubscript{\textcolor{gray}{[$-$2.30,$-$0.30]}} \\
\cmidrule(l){2-7}
    & \multirow{3}{*}{\texttt{ED-Decomp}} & \multirow{3}{*}{PPG (bedside)} & F1-Macro & \underline{+0.19}\textsubscript{\textcolor{gray}{[+0.02,+0.38]}} & +0.05\textsubscript{\textcolor{gray}{[$-$1.47,+1.54]}} & \textbf{+1.40}\textsubscript{\textcolor{gray}{[$-$0.42,+3.11]}} \\
    & & & Bal.\ acc. & \underline{+0.23}\textsubscript{\textcolor{gray}{[+0.02,+0.41]}} & +0.00\textsubscript{\textcolor{gray}{[$-$0.79,+0.75]}} & \textbf{+1.08}\textsubscript{\textcolor{gray}{[$-$0.26,+2.35]}} \\
    & & & Acc. & \textbf{+0.68}\textsubscript{\textcolor{gray}{[+0.10,+1.35]}} & \underline{+0.38}\textsubscript{\textcolor{gray}{[$-$0.27,+0.96]}} & $-$0.12\textsubscript{\textcolor{gray}{[$-$0.96,+0.86]}} \\
\cmidrule(l){2-7}
    & \multirow{3}{*}{\texttt{ED-Revisit}} & \multirow{3}{*}{ECG (bedside)} & F1-Macro & +0.00\textsubscript{\textcolor{gray}{[$-$1.34,+1.04]}} & \underline{+0.13}\textsubscript{\textcolor{gray}{[$-$1.24,+1.57]}} & \textbf{+0.97}\textsubscript{\textcolor{gray}{[$-$0.86,+2.93]}} \\
    & & & Bal.\ acc. & \underline{$-$0.09}\textsubscript{\textcolor{gray}{[$-$0.84,+0.49]}} & $-$0.94\textsubscript{\textcolor{gray}{[$-$3.31,+1.26]}} & \textbf{+0.38}\textsubscript{\textcolor{gray}{[$-$0.96,+1.72]}} \\
    & & & Acc. & \underline{$-$0.32}\textsubscript{\textcolor{gray}{[$-$0.71,+0.10]}} & \textbf{+13.52}\textsubscript{\textcolor{gray}{[+12.26,+14.79]}} & $-$0.47\textsubscript{\textcolor{gray}{[$-$1.29,+0.22]}} \\
\cmidrule(l){2-7}
    & \multirow{3}{*}{\texttt{ED-Revisit}} & \multirow{3}{*}{PPG (bedside)} & F1-Macro & \underline{$-$0.04}\textsubscript{\textcolor{gray}{[$-$1.42,+1.34]}} & $-$0.52\textsubscript{\textcolor{gray}{[$-$1.76,+0.72]}} & \textbf{+1.26}\textsubscript{\textcolor{gray}{[$-$0.20,+2.80]}} \\
    & & & Bal.\ acc. & \underline{$-$0.13}\textsubscript{\textcolor{gray}{[$-$0.93,+0.60]}} & $-$1.57\textsubscript{\textcolor{gray}{[$-$3.69,+0.42]}} & \textbf{+0.86}\textsubscript{\textcolor{gray}{[$-$0.25,+2.11]}} \\
    & & & Acc. & \underline{$-$0.46}\textsubscript{\textcolor{gray}{[$-$0.91,+0.00]}} & \textbf{+1.71}\textsubscript{\textcolor{gray}{[+0.39,+3.08]}} & $-$0.46\textsubscript{\textcolor{gray}{[$-$1.17,+0.32]}} \\
\bottomrule
\end{tabular}%
}
\vspace{2pt}
\caption{\dlift\ (\texttt{+wav} $-$ \texttt{EHR}) macro-F1 and balanced accuracy, with 95\% paired-bootstrap CIs (B{=}200) as grey subscripts. In each row, the \textbf{best} value is bold and second best is \underline{underlined.}}
\label{tab:cond-lift-table}
\end{table}

\newcommand{\rawagg}[6]{%
\begin{tabular}{@{}c@{}}
F1: #1$\rightarrow$#2\\
BA: #3$\rightarrow$#4\\
Acc: #5$\rightarrow$#6
\end{tabular}}

\begin{table}[h]
\centering
\setlength{\tabcolsep}{1.2pt}
\renewcommand{\arraystretch}{1.5}
\caption{Binary-task class diagnostics for conditional modality lift. Each model cell reports $\Delta$Sensitivity / $\Delta$Specificity / $\Delta F1_{\mathrm{event}}$ in percentage points after adding the physiological signal to the fixed EHR-only input.}
\label{tab:app-binary-class-diagnostics}
\resizebox{\textwidth}{!}{%
\begin{tabular}{@{}llllcccc@{}}
\toprule
\textbf{Dataset} & \textbf{Task} & \textbf{Signal} & \textbf{Event label} & \textbf{QoQ-Med} & \textbf{MedGemma-4B} & \textbf{MedGemma-27B} & \textbf{Supervised} \\
\midrule
MIMIC-IV & \texttt{48-IHM} & ECG (12-lead) & no & \diagbin{-12.1}{+15.1}{+1.4} & \diagbin{+12.4}{-5.8}{+2.4} & \diagbin{0.0}{+10.4}{+0.8} & \diagbin{-16.9}{+14.4}{+3.8} \\
MCMED & \texttt{ED-Decomp} & ECG (bedside) & decomp & \diagbin{0.0}{+0.7}{0.0} & \diagbin{-0.4}{0.0}{-0.8} & \diagbin{-1.1}{+0.1}{-1.5} & \diagbin{-5.1}{+11.0}{+2.7} \\
MCMED & \texttt{ED-Decomp} & PPG (bedside) & decomp & \diagbin{0.0}{+0.5}{0.0} & \diagbin{-0.1}{+0.1}{-0.2} & \diagbin{+3.1}{-0.9}{+2.9} & \diagbin{+6.9}{+2.4}{+4.3} \\
\bottomrule
\end{tabular}%
}
\end{table}

\begin{table}[h]
\centering
\setlength{\tabcolsep}{4.2pt}
\caption{Multiclass class diagnostics for conditional modality lift. Each model cell reports $\Delta$Recall / $\Delta F1_c$ in percentage points after adding the physiological signal to the fixed EHR-only input.}
\label{tab:app-multiclass-class-diagnostics}
\resizebox{\textwidth}{!}{%
\begin{tabular}{@{}llllcccc@{}}
\toprule
\textbf{Dataset} & \textbf{Task} & \textbf{Signal} & \textbf{Class} & \textbf{QoQ-Med} & \textbf{MedGemma-4B} & \textbf{MedGemma-27B} & \textbf{Supervised} \\
\midrule
MIMIC-IV & \texttt{LOS} & ECG (12-lead) & A & \diagmulti{+15.7}{+8.7} & \diagmulti{-1.6}{-3.1} & \diagmulti{-5.8}{-7.8} & \diagmulti{-1.9}{-1.2} \\
 &  &  & B & \diagmulti{-9.1}{-3.8} & \diagmulti{-8.3}{-1.4} & \diagmulti{-1.9}{-1.2} & \diagmulti{+6.4}{+3.6} \\
 &  &  & C & \diagmulti{-0.9}{-1.3} & \diagmulti{+3.4}{+4.9} & \diagmulti{+0.6}{+0.8} & \diagmulti{+0.9}{+1.7} \\
 &  &  & D & \diagmulti{-2.8}{+1.4} & \diagmulti{0.0}{0.0} & \diagmulti{+1.3}{-2.1} & \diagmulti{-0.9}{+0.8} \\
MCMED & \texttt{ED-Revisit} & ECG (bedside) & 0--3d & \diagmulti{0.0}{0.0} & \diagmulti{0.0}{+0.5} & \diagmulti{-1.9}{-2.5} & \diagmulti{-7.6}{+0.3} \\
 &  &  & 3--7d & \diagmulti{+1.0}{+1.9} & \diagmulti{-19.8}{-7.5} & \diagmulti{+4.0}{+6.7} & \diagmulti{-8.9}{-1.1} \\
 &  &  & 7--14d & \diagmulti{-1.0}{-1.7} & \diagmulti{0.0}{0.0} & \diagmulti{0.0}{0.0} & \diagmulti{+5.8}{+6.4} \\
 &  &  & none & \diagmulti{-0.4}{-0.2} & \diagmulti{+16.1}{+7.6} & \diagmulti{-0.6}{-0.3} & \diagmulti{+14.1}{+12.0} \\
MCMED & \texttt{ED-Revisit} & PPG (bedside) & 0--3d & \diagmulti{0.0}{0.0} & \diagmulti{-0.6}{-0.9} & \diagmulti{+3.2}{+3.7} & \diagmulti{-14.6}{+2.3} \\
 &  &  & 3--7d & \diagmulti{0.0}{0.0} & \diagmulti{-7.9}{-2.3} & \diagmulti{+1.0}{+1.6} & \diagmulti{-6.9}{-2.9} \\
 &  &  & 7--14d & \diagmulti{0.0}{+0.1} & \diagmulti{0.0}{0.0} & \diagmulti{0.0}{0.0} & \diagmulti{+2.9}{-0.6} \\
 &  &  & none & \diagmulti{-0.5}{-0.3} & \diagmulti{+2.3}{+1.1} & \diagmulti{-0.7}{-0.3} & \diagmulti{+15.8}{+13.3} \\
\bottomrule
\end{tabular}%
}
\end{table}

\section{Subgroup Conditional Modality Lift}
\label{app:subgroup_tables}
\begin{table}[H]
\centering
\scriptsize
\setlength{\tabcolsep}{4.2pt}
\renewcommand{\arraystretch}{1.18}
\caption{Per-subgroup conditional modality lift in macro-F1 for MIMIC-IV \texttt{48-IHM} .}
\label{tab:subgroup-48ihm}
\resizebox{\textwidth}{!}{%
\begin{tabular}{@{}llcccccc@{}}
\toprule
\textbf{Axis} & \textbf{Level} & \multicolumn{2}{c}{\textbf{QoQ-Omni}} & \multicolumn{2}{c}{\textbf{MedGemma-4B}} & \multicolumn{2}{c}{\textbf{MedGemma-27B}} \\
\cmidrule(lr){3-4}\cmidrule(lr){5-6}\cmidrule(lr){7-8}
 &  & \textbf{F1-Macro} & $\boldsymbol{\Delta}_{\mathrm{lift}}$ & \textbf{F1-Macro} & $\boldsymbol{\Delta}_{\mathrm{lift}}$ & \textbf{F1-Macro} & $\boldsymbol{\Delta}_{\mathrm{lift}}$ \\
\midrule
Charlson CCI & 0 & \scorecell{40.01}{43.05} & \subcell{+3.04}{+0.51}{+5.65} & \scorecell{53.68}{50.19} & \subcell{-3.49}{-9.24}{+1.76} & \scorecell{37.26}{44.69} & \subcell{+7.43}{+4.54}{+10.13} \\
 & 1-2 & \scorecell{42.20}{46.17} & \subcell{+3.97}{+1.82}{+6.00} & \scorecell{52.30}{53.30} & \subcell{+1.00}{-2.80}{+4.75} & \scorecell{39.13}{43.86} & \subcell{+4.74}{+2.83}{+6.72} \\
 & 3-5 & \scorecell{41.85}{49.53} & \subcell{+7.68}{+5.37}{+10.09} & \scorecell{53.22}{53.47} & \subcell{+0.25}{-3.50}{+4.12} & \scorecell{33.62}{37.79} & \subcell{+4.17}{+2.15}{+5.97} \\
 & 6+ & \scorecell{43.64}{53.71} & \subcell{+10.07}{+6.17}{+14.10} & \scorecell{54.35}{53.43} & \subcell{-0.91}{-4.95}{+3.43} & \scorecell{34.67}{40.47} & \subcell{+5.80}{+2.90}{+8.88} \\
\addlinespace[2pt]
Age & \textless{}40 & \scorecell{41.06}{45.29} & \subcell{+4.24}{+0.42}{+8.64} & \scorecell{52.45}{47.41} & \subcell{-5.04}{-13.31}{+2.64} & \scorecell{36.33}{43.35} & \subcell{+7.02}{+3.89}{+10.29} \\
 & 40-60 & \scorecell{40.73}{45.57} & \subcell{+4.84}{+2.34}{+7.26} & \scorecell{54.41}{50.78} & \subcell{-3.64}{-8.49}{+1.42} & \scorecell{36.03}{41.97} & \subcell{+5.94}{+4.09}{+7.87} \\
 & 60-75 & \scorecell{44.84}{50.81} & \subcell{+5.96}{+3.28}{+8.48} & \scorecell{52.48}{53.21} & \subcell{+0.74}{-2.61}{+4.18} & \scorecell{38.78}{42.77} & \subcell{+3.99}{+1.82}{+5.98} \\
 & 75+ & \scorecell{41.13}{48.88} & \subcell{+7.75}{+5.10}{+10.51} & \scorecell{54.07}{54.56} & \subcell{+0.49}{-2.77}{+4.06} & \scorecell{34.86}{40.27} & \subcell{+5.41}{+3.13}{+7.66} \\
\addlinespace[2pt]
Sex & F & \scorecell{42.42}{48.19} & \subcell{+5.76}{+3.88}{+7.70} & \scorecell{55.70}{54.03} & \subcell{-1.67}{-4.91}{+1.45} & \scorecell{38.37}{44.35} & \subcell{+5.98}{+4.32}{+7.57} \\
 & M & \scorecell{42.49}{48.99} & \subcell{+6.50}{+4.60}{+8.48} & \scorecell{51.70}{52.70} & \subcell{+1.01}{-1.68}{+3.81} & \scorecell{35.89}{40.61} & \subcell{+4.72}{+3.08}{+6.40} \\
\addlinespace[2pt]
Race & WHITE & \scorecell{42.65}{48.63} & \subcell{+5.99}{+4.29}{+7.64} & \scorecell{54.28}{53.08} & \subcell{-1.21}{-3.63}{+1.31} & \scorecell{37.07}{42.68} & \subcell{+5.62}{+4.26}{+6.96} \\
 & BLACK & \scorecell{40.59}{45.75} & \subcell{+5.16}{+2.18}{+8.11} & \scorecell{53.37}{52.17} & \subcell{-1.20}{-6.85}{+4.57} & \scorecell{34.52}{39.47} & \subcell{+4.95}{+2.52}{+7.67} \\
 & ASIAN & \scorecell{46.39}{53.40} & \subcell{+7.01}{-1.31}{+15.48} & \scorecell{50.97}{57.06} & \subcell{+6.09}{-5.40}{+18.29} & \scorecell{36.97}{42.00} & \subcell{+5.03}{-2.07}{+12.28} \\
 & HISPANIC & \scorecell{38.33}{45.00} & \subcell{+6.68}{+1.75}{+11.79} & \scorecell{51.77}{47.66} & \subcell{-4.11}{-12.89}{+4.47} & \scorecell{34.80}{39.99} & \subcell{+5.19}{+1.03}{+9.15} \\
 & OTHER/UNKNOWN & \scorecell{45.02}{53.30} & \subcell{+8.28}{+2.00}{+14.98} & \scorecell{49.91}{58.20} & \subcell{+8.29}{+0.60}{+15.42} & \scorecell{45.44}{48.63} & \subcell{+3.20}{-1.23}{+7.83} \\
\addlinespace[2pt]
Anchor year & 2008 - 2010 & \scorecell{40.42}{47.25} & \subcell{+6.83}{+5.01}{+8.58} & \scorecell{51.21}{51.26} & \subcell{+0.05}{-3.10}{+3.27} & \scorecell{35.64}{41.31} & \subcell{+5.67}{+4.07}{+7.47} \\
 & 2011 - 2013 & \scorecell{42.49}{48.80} & \subcell{+6.31}{+3.49}{+8.97} & \scorecell{52.21}{53.09} & \subcell{+0.88}{-3.01}{+4.68} & \scorecell{38.02}{43.07} & \subcell{+5.06}{+2.80}{+7.28} \\
 & 2014 - 2016 & \scorecell{45.99}{49.75} & \subcell{+3.76}{+0.78}{+6.79} & \scorecell{57.83}{56.05} & \subcell{-1.78}{-5.86}{+2.52} & \scorecell{38.08}{43.27} & \subcell{+5.19}{+2.50}{+7.61} \\
\addlinespace[2pt]
Admission type & Emergency & \scorecell{41.61}{43.12} & \subcell{+1.51}{-2.72}{+5.88} & \scorecell{55.50}{54.11} & \subcell{-1.39}{-9.07}{+6.17} & \scorecell{28.69}{40.55} & \subcell{+11.87}{+7.70}{+15.72} \\
 & Observation & \scorecell{40.74}{47.41} & \subcell{+6.67}{+3.90}{+9.69} & \scorecell{54.80}{51.34} & \subcell{-3.46}{-8.11}{+1.24} & \scorecell{38.53}{43.59} & \subcell{+5.06}{+2.63}{+7.37} \\
 & Other & \scorecell{42.90}{49.20} & \subcell{+6.31}{+4.55}{+7.96} & \scorecell{53.48}{53.85} & \subcell{+0.37}{-2.10}{+2.85} & \scorecell{37.50}{42.38} & \subcell{+4.87}{+3.43}{+6.25} \\
\addlinespace[2pt]
Insurance & Medicaid & \scorecell{43.20}{46.16} & \subcell{+2.96}{-0.44}{+6.27} & \scorecell{51.02}{48.94} & \subcell{-2.08}{-8.13}{+3.35} & \scorecell{36.63}{41.67} & \subcell{+5.04}{+2.44}{+7.54} \\
 & Medicare & \scorecell{42.42}{48.97} & \subcell{+6.55}{+4.81}{+8.45} & \scorecell{53.14}{53.94} & \subcell{+0.80}{-1.91}{+3.54} & \scorecell{36.24}{41.70} & \subcell{+5.46}{+4.02}{+6.98} \\
 & Other & \scorecell{41.58}{48.55} & \subcell{+6.97}{+3.72}{+9.80} & \scorecell{56.82}{53.49} & \subcell{-3.33}{-8.63}{+1.39} & \scorecell{39.08}{44.33} & \subcell{+5.25}{+2.91}{+7.77} \\
\addlinespace[2pt]
\bottomrule
\end{tabular}%
}
\end{table}

\begin{table}[H]
\centering
\scriptsize
\setlength{\tabcolsep}{4.2pt}
\renewcommand{\arraystretch}{1.18}
\caption{Per-subgroup conditional modality lift in macro-F1 for MIMIC-IV \texttt{LOS} .}
\label{tab:subgroup-los}
\resizebox{\textwidth}{!}{%
\begin{tabular}{@{}llcccccc@{}}
\toprule
\textbf{Axis} & \textbf{Level} & \multicolumn{2}{c}{\textbf{QoQ-Omni}} & \multicolumn{2}{c}{\textbf{MedGemma-4B}} & \multicolumn{2}{c}{\textbf{MedGemma-27B}} \\
\cmidrule(lr){3-4}\cmidrule(lr){5-6}\cmidrule(lr){7-8}
 &  & \textbf{F1-Macro} & $\boldsymbol{\Delta}_{\mathrm{lift}}$ & \textbf{F1-Macro} & $\boldsymbol{\Delta}_{\mathrm{lift}}$ & \textbf{F1-Macro} & $\boldsymbol{\Delta}_{\mathrm{lift}}$ \\
\midrule
Charlson CCI & 1-2 & \scorecell{26.81}{28.27} & \subcell{+1.46}{-2.36}{+5.31} & \scorecell{13.30}{11.11} & \subcell{-2.19}{-3.11}{-1.20} & \scorecell{26.19}{23.06} & \subcell{-3.13}{-6.31}{+0.31} \\
 & 3-5 & \scorecell{24.11}{25.10} & \subcell{+0.98}{-1.09}{+3.15} & \scorecell{13.08}{13.80} & \subcell{+0.72}{-0.38}{+1.72} & \scorecell{23.37}{20.10} & \subcell{-3.27}{-5.42}{-1.24} \\
 & 6+ & \scorecell{24.51}{26.50} & \subcell{+1.99}{-1.26}{+5.30} & \scorecell{11.67}{12.66} & \subcell{+0.99}{-0.28}{+2.34} & \scorecell{22.52}{22.61} & \subcell{+0.09}{-2.84}{+3.03} \\
\addlinespace[2pt]
Age & \textless{}40 & \scorecell{27.28}{25.82} & \subcell{-1.46}{-9.42}{+6.11} & \scorecell{11.80}{9.46} & \subcell{-2.34}{-4.94}{-0.25} & \scorecell{25.22}{19.21} & \subcell{-6.01}{-13.32}{+0.73} \\
 & 40-60 & \scorecell{24.70}{26.09} & \subcell{+1.39}{-1.68}{+4.61} & \scorecell{12.59}{13.15} & \subcell{+0.56}{-1.24}{+2.45} & \scorecell{26.38}{24.52} & \subcell{-1.86}{-5.19}{+1.43} \\
 & 60-75 & \scorecell{27.30}{26.19} & \subcell{-1.11}{-3.77}{+1.84} & \scorecell{12.90}{11.92} & \subcell{-0.98}{-2.19}{+0.28} & \scorecell{25.65}{23.69} & \subcell{-1.96}{-4.39}{+0.69} \\
 & 75+ & \scorecell{23.12}{26.41} & \subcell{+3.29}{+1.02}{+5.79} & \scorecell{13.06}{13.94} & \subcell{+0.89}{-0.25}{+2.19} & \scorecell{22.44}{19.53} & \subcell{-2.91}{-5.15}{-0.58} \\
\addlinespace[2pt]
Sex & F & \scorecell{25.45}{25.22} & \subcell{-0.23}{-2.67}{+1.99} & \scorecell{13.11}{13.13} & \subcell{+0.02}{-0.97}{+1.12} & \scorecell{24.75}{21.25} & \subcell{-3.50}{-5.67}{-1.28} \\
 & M & \scorecell{24.94}{27.40} & \subcell{+2.47}{+0.14}{+4.71} & \scorecell{12.63}{12.82} & \subcell{+0.19}{-0.82}{+1.22} & \scorecell{24.33}{22.54} & \subcell{-1.79}{-3.99}{+0.27} \\
\addlinespace[2pt]
Race & WHITE & \scorecell{24.14}{25.98} & \subcell{+1.83}{-0.10}{+3.91} & \scorecell{12.77}{12.37} & \subcell{-0.40}{-1.20}{+0.37} & \scorecell{24.82}{21.76} & \subcell{-3.06}{-4.80}{-1.41} \\
 & BLACK & \scorecell{27.30}{25.68} & \subcell{-1.61}{-5.13}{+1.72} & \scorecell{12.62}{13.14} & \subcell{+0.52}{-1.39}{+2.42} & \scorecell{22.86}{22.03} & \subcell{-0.83}{-4.27}{+2.27} \\
 & ASIAN & \scorecell{28.44}{29.02} & \subcell{+0.57}{-10.33}{+11.52} & \scorecell{13.26}{12.40} & \subcell{-0.87}{-3.82}{+1.36} & \scorecell{22.95}{21.64} & \subcell{-1.32}{-12.55}{+9.24} \\
 & HISPANIC & \scorecell{25.24}{28.06} & \subcell{+2.82}{-3.27}{+8.46} & \scorecell{15.79}{19.27} & \subcell{+3.48}{-2.85}{+9.25} & \scorecell{22.48}{21.48} & \subcell{-1.00}{-5.38}{+3.04} \\
 & OTHER/UNKNOWN & \scorecell{27.25}{30.60} & \subcell{+3.35}{-3.21}{+9.79} & \scorecell{11.90}{15.75} & \subcell{+3.84}{+0.62}{+7.54} & \scorecell{28.74}{23.63} & \subcell{-5.11}{-12.58}{+2.25} \\
\addlinespace[2pt]
Anchor year & 2008 - 2010 & \scorecell{24.17}{25.72} & \subcell{+1.55}{-0.49}{+3.59} & \scorecell{12.82}{13.00} & \subcell{+0.18}{-0.96}{+1.25} & \scorecell{22.92}{21.17} & \subcell{-1.75}{-3.53}{+0.17} \\
 & 2011 - 2013 & \scorecell{26.21}{27.02} & \subcell{+0.82}{-2.53}{+4.17} & \scorecell{12.94}{12.33} & \subcell{-0.61}{-1.89}{+0.72} & \scorecell{25.57}{20.51} & \subcell{-5.06}{-7.90}{-2.16} \\
 & 2014 - 2016 & \scorecell{26.33}{26.92} & \subcell{+0.58}{-2.70}{+4.12} & \scorecell{12.83}{13.33} & \subcell{+0.50}{-0.98}{+2.35} & \scorecell{26.88}{26.20} & \subcell{-0.68}{-4.59}{+3.51} \\
\addlinespace[2pt]
Admission type & Emergency & \scorecell{21.39}{24.58} & \subcell{+3.19}{-1.76}{+8.37} & \scorecell{12.37}{14.54} & \subcell{+2.17}{-0.03}{+4.67} & \scorecell{22.25}{19.24} & \subcell{-3.01}{-7.88}{+2.35} \\
 & Observation & \scorecell{24.92}{25.91} & \subcell{+0.99}{-2.03}{+3.94} & \scorecell{12.84}{14.18} & \subcell{+1.34}{-0.53}{+3.50} & \scorecell{30.76}{26.37} & \subcell{-4.39}{-7.79}{-0.93} \\
 & Other & \scorecell{25.43}{26.47} & \subcell{+1.04}{-1.02}{+3.04} & \scorecell{12.73}{12.22} & \subcell{-0.51}{-1.35}{+0.31} & \scorecell{21.82}{19.90} & \subcell{-1.93}{-3.63}{-0.09} \\
\addlinespace[2pt]
Insurance & Medicaid & \scorecell{25.26}{24.15} & \subcell{-1.11}{-5.05}{+2.94} & \scorecell{13.67}{13.27} & \subcell{-0.40}{-2.96}{+2.23} & \scorecell{24.66}{24.64} & \subcell{-0.03}{-4.18}{+4.55} \\
 & Medicare & \scorecell{24.82}{26.28} & \subcell{+1.45}{-0.59}{+3.47} & \scorecell{12.76}{12.93} & \subcell{+0.16}{-0.66}{+1.02} & \scorecell{23.81}{20.81} & \subcell{-2.99}{-4.68}{-1.27} \\
 & Other & \scorecell{26.50}{28.57} & \subcell{+2.07}{-2.26}{+6.02} & \scorecell{12.78}{13.01} & \subcell{+0.22}{-1.68}{+2.29} & \scorecell{28.73}{26.55} & \subcell{-2.17}{-6.24}{+2.20} \\
\addlinespace[2pt]
\bottomrule
\end{tabular}%
}
\end{table}

\begin{table}[H]
\centering
\scriptsize
\setlength{\tabcolsep}{4.2pt}
\renewcommand{\arraystretch}{1.18}
\caption{Per-subgroup conditional modality lift in macro-F1 for MC-MED \texttt{ED-Decomp} with ECG (bedside).}
\label{tab:subgroup-eddecomp-ecg}
\resizebox{\textwidth}{!}{%
\begin{tabular}{@{}llcccccc@{}}
\toprule
\textbf{Axis} & \textbf{Level} & \multicolumn{2}{c}{\textbf{QoQ-Omni}} & \multicolumn{2}{c}{\textbf{MedGemma-4B}} & \multicolumn{2}{c}{\textbf{MedGemma-27B}} \\
\cmidrule(lr){3-4}\cmidrule(lr){5-6}\cmidrule(lr){7-8}
 &  & \textbf{F1-Macro} & $\boldsymbol{\Delta}_{\mathrm{lift}}$ & \textbf{F1-Macro} & $\boldsymbol{\Delta}_{\mathrm{lift}}$ & \textbf{F1-Macro} & $\boldsymbol{\Delta}_{\mathrm{lift}}$ \\
\midrule
Charlson CCI & 0 & \scorecell{46.16}{46.18} & \subcell{+0.03}{-0.05}{+0.14} & \scorecell{47.54}{46.90} & \subcell{-0.64}{-2.06}{+0.13} & \scorecell{51.27}{51.45} & \subcell{+0.18}{-2.90}{+3.58} \\
 & 1-2 & \scorecell{45.19}{45.32} & \subcell{+0.14}{+0.00}{+0.31} & \scorecell{49.47}{49.47} & \subcell{+0.00}{-3.86}{+3.62} & \scorecell{57.90}{58.57} & \subcell{+0.66}{-2.95}{+4.74} \\
 & 3-5 & \scorecell{44.91}{44.91} & \subcell{+0.00}{-0.24}{+0.24} & \scorecell{50.58}{47.86} & \subcell{-2.72}{-6.54}{+0.15} & \scorecell{57.41}{54.68} & \subcell{-2.73}{-8.26}{+2.42} \\
 & 6+ & \scorecell{44.53}{44.79} & \subcell{+0.27}{+0.00}{+0.69} & \scorecell{46.99}{46.63} & \subcell{-0.36}{-1.19}{+0.27} & \scorecell{50.88}{50.08} & \subcell{-0.80}{-7.05}{+5.50} \\
\addlinespace[2pt]
Age & \textless{}40 & -- & -- & \scorecell{48.58}{45.81} & \subcell{-2.77}{-7.02}{+0.06} & \scorecell{52.06}{53.94} & \subcell{+1.88}{-3.51}{+7.22} \\
 & 40-60 & \scorecell{45.68}{45.80} & \subcell{+0.13}{+0.00}{+0.29} & \scorecell{48.39}{48.35} & \subcell{-0.04}{-2.47}{+2.58} & \scorecell{56.50}{53.89} & \subcell{-2.60}{-6.25}{+0.73} \\
 & 60-75 & \scorecell{45.14}{45.20} & \subcell{+0.05}{-0.11}{+0.26} & \scorecell{49.30}{48.30} & \subcell{-1.00}{-4.56}{+2.14} & \scorecell{55.64}{56.55} & \subcell{+0.92}{-2.93}{+5.00} \\
 & 75+ & \scorecell{45.48}{45.58} & \subcell{+0.10}{-0.10}{+0.30} & \scorecell{48.59}{48.59} & \subcell{+0.00}{-2.61}{+2.93} & \scorecell{53.00}{52.99} & \subcell{-0.02}{-4.65}{+4.56} \\
\addlinespace[2pt]
Sex & F & \scorecell{45.15}{45.28} & \subcell{+0.13}{+0.00}{+0.28} & \scorecell{46.70}{47.17} & \subcell{+0.47}{-1.70}{+2.61} & \scorecell{51.80}{52.13} & \subcell{+0.33}{-2.06}{+2.79} \\
 & M & \scorecell{45.89}{45.91} & \subcell{+0.02}{-0.05}{+0.12} & \scorecell{50.89}{48.73} & \subcell{-2.15}{-4.20}{-0.41} & \scorecell{57.42}{56.52} & \subcell{-0.90}{-4.21}{+2.63} \\
\addlinespace[2pt]
Race & WHITE & \scorecell{44.94}{45.06} & \subcell{+0.12}{+0.03}{+0.25} & \scorecell{48.33}{48.38} & \subcell{+0.05}{-2.16}{+2.00} & \scorecell{53.61}{52.35} & \subcell{-1.26}{-4.37}{+1.75} \\
 & BLACK & \scorecell{46.63}{46.38} & \subcell{-0.26}{-0.80}{+0.00} & -- & -- & -- & -- \\
 & ASIAN & \scorecell{46.47}{46.53} & \subcell{+0.07}{-0.13}{+0.32} & \scorecell{48.35}{48.44} & \subcell{+0.10}{+0.00}{+0.36} & \scorecell{52.59}{52.27} & \subcell{-0.32}{-4.37}{+3.78} \\
 & OTHER/UNKNOWN & \scorecell{45.57}{45.64} & \subcell{+0.07}{+0.00}{+0.19} & \scorecell{48.66}{46.24} & \subcell{-2.42}{-5.88}{+0.68} & \scorecell{56.59}{57.65} & \subcell{+1.06}{-2.92}{+5.50} \\
\addlinespace[2pt]
Insurance & Medicaid & -- & -- & \scorecell{48.98}{45.54} & \subcell{-3.44}{-8.62}{+0.00} & \scorecell{57.91}{54.53} & \subcell{-3.38}{-9.70}{+3.86} \\
 & Medicare & \scorecell{44.87}{44.91} & \subcell{+0.04}{-0.08}{+0.16} & \scorecell{47.65}{47.70} & \subcell{+0.06}{-1.81}{+2.06} & \scorecell{52.69}{52.06} & \subcell{-0.64}{-3.93}{+2.64} \\
 & Other & \scorecell{46.24}{46.35} & \subcell{+0.11}{+0.00}{+0.23} & \scorecell{49.15}{48.57} & \subcell{-0.58}{-2.99}{+1.73} & \scorecell{54.53}{55.77} & \subcell{+1.24}{-1.49}{+4.24} \\
\addlinespace[2pt]
Arrival mode & EMS & \scorecell{44.63}{44.74} & \subcell{+0.11}{+0.00}{+0.29} & \scorecell{47.30}{47.46} & \subcell{+0.16}{-2.59}{+2.62} & \scorecell{53.12}{55.71} & \subcell{+2.59}{-1.28}{+6.87} \\
 & Self & \scorecell{45.79}{45.85} & \subcell{+0.07}{-0.02}{+0.16} & \scorecell{49.18}{48.03} & \subcell{-1.14}{-3.14}{+0.74} & \scorecell{55.04}{53.77} & \subcell{-1.27}{-3.68}{+1.33} \\
\addlinespace[2pt]
Triage acuity & 2 & \scorecell{45.04}{45.09} & \subcell{+0.05}{-0.10}{+0.24} & \scorecell{48.44}{46.89} & \subcell{-1.55}{-4.23}{+0.29} & \scorecell{53.25}{53.75} & \subcell{+0.50}{-3.31}{+4.18} \\
 & 3 & \scorecell{45.73}{45.82} & \subcell{+0.09}{+0.00}{+0.19} & \scorecell{48.56}{48.08} & \subcell{-0.48}{-2.53}{+1.45} & \scorecell{55.18}{54.34} & \subcell{-0.84}{-3.51}{+1.89} \\
\addlinespace[2pt]
\bottomrule
\end{tabular}%
}
\end{table}

\begin{table}[H]
\centering
\scriptsize
\setlength{\tabcolsep}{4.2pt}
\renewcommand{\arraystretch}{1.18}
\caption{Per-subgroup conditional modality lift in macro-F1 for MC-MED \texttt{ED-Revisit} with ECG (bedside).}
\label{tab:subgroup-edrev-ecg}
\resizebox{\textwidth}{!}{%
\begin{tabular}{@{}llcccccc@{}}
\toprule
\textbf{Axis} & \textbf{Level} & \multicolumn{2}{c}{\textbf{QoQ-Omni}} & \multicolumn{2}{c}{\textbf{MedGemma-4B}} & \multicolumn{2}{c}{\textbf{MedGemma-27B}} \\
\cmidrule(lr){3-4}\cmidrule(lr){5-6}\cmidrule(lr){7-8}
 &  & \textbf{F1-Macro} & $\boldsymbol{\Delta}_{\mathrm{lift}}$ & \textbf{F1-Macro} & $\boldsymbol{\Delta}_{\mathrm{lift}}$ & \textbf{F1-Macro} & $\boldsymbol{\Delta}_{\mathrm{lift}}$ \\
\midrule
Charlson CCI & 0 & \scorecell{23.93}{23.89} & \subcell{-0.04}{-0.09}{+0.01} & \scorecell{24.25}{24.51} & \subcell{+0.25}{-1.69}{+2.37} & \scorecell{24.62}{24.03} & \subcell{-0.59}{-2.30}{+0.79} \\
 & 1-2 & \scorecell{22.55}{22.53} & \subcell{-0.02}{-0.17}{+0.13} & \scorecell{23.39}{22.58} & \subcell{-0.80}{-2.78}{+1.12} & \scorecell{25.03}{26.88} & \subcell{+1.85}{-2.00}{+5.99} \\
 & 3-5 & \scorecell{24.33}{25.22} & \subcell{+0.88}{-6.48}{+8.44} & \scorecell{20.60}{24.46} & \subcell{+3.86}{-0.38}{+8.91} & \scorecell{23.70}{29.90} & \subcell{+6.20}{-1.34}{+13.86} \\
\addlinespace[2pt]
Age & \textless{}40 & \scorecell{23.51}{23.47} & \subcell{-0.04}{-0.11}{+0.04} & \scorecell{24.39}{24.41} & \subcell{+0.01}{-2.54}{+2.63} & \scorecell{24.01}{26.35} & \subcell{+2.34}{-0.10}{+5.46} \\
 & 40-60 & \scorecell{25.09}{23.66} & \subcell{-1.43}{-4.33}{-0.02} & \scorecell{23.81}{23.68} & \subcell{-0.13}{-2.05}{+1.43} & \scorecell{25.52}{27.34} & \subcell{+1.83}{-2.24}{+6.40} \\
 & 60-75 & \scorecell{22.53}{22.56} & \subcell{+0.03}{-0.13}{+0.18} & \scorecell{22.42}{23.50} & \subcell{+1.08}{-1.52}{+3.94} & \scorecell{24.61}{23.34} & \subcell{-1.28}{-3.73}{+1.09} \\
 & 75+ & \scorecell{23.39}{26.80} & \subcell{+3.41}{-0.44}{+9.66} & \scorecell{23.54}{23.23} & \subcell{-0.31}{-3.57}{+2.75} & \scorecell{26.33}{26.03} & \subcell{-0.30}{-7.64}{+7.43} \\
\addlinespace[2pt]
Sex & F & \scorecell{24.47}{23.55} & \subcell{-0.93}{-2.92}{+0.02} & \scorecell{23.19}{23.59} & \subcell{+0.40}{-0.76}{+1.42} & \scorecell{25.07}{26.98} & \subcell{+1.91}{-1.30}{+5.29} \\
 & M & \scorecell{23.16}{24.10} & \subcell{+0.94}{-0.14}{+2.97} & \scorecell{24.46}{24.25} & \subcell{-0.22}{-2.67}{+2.06} & \scorecell{24.89}{24.88} & \subcell{-0.01}{-2.03}{+2.31} \\
\addlinespace[2pt]
Race & WHITE & \scorecell{23.33}{24.72} & \subcell{+1.39}{-0.13}{+4.51} & \scorecell{23.59}{23.31} & \subcell{-0.28}{-2.29}{+1.45} & \scorecell{25.50}{25.40} & \subcell{-0.10}{-3.01}{+3.32} \\
 & BLACK & \scorecell{26.26}{22.43} & \subcell{-3.83}{-11.06}{+0.24} & \scorecell{21.91}{22.49} & \subcell{+0.58}{-2.58}{+2.69} & \scorecell{30.35}{34.29} & \subcell{+3.94}{-7.21}{+15.22} \\
 & ASIAN & \scorecell{23.65}{23.65} & \subcell{+0.00}{-0.12}{+0.12} & \scorecell{25.48}{25.87} & \subcell{+0.39}{-4.02}{+5.75} & \scorecell{23.24}{23.23} & \subcell{-0.00}{-0.23}{+0.21} \\
 & OTHER/UNKNOWN & \scorecell{23.51}{23.41} & \subcell{-0.10}{-0.19}{-0.02} & \scorecell{23.58}{24.14} & \subcell{+0.55}{-1.65}{+2.69} & \scorecell{24.25}{25.37} & \subcell{+1.12}{-1.19}{+4.40} \\
\addlinespace[2pt]
Insurance & Medicaid & \scorecell{23.15}{23.04} & \subcell{-0.11}{-0.27}{+0.04} & \scorecell{25.44}{23.08} & \subcell{-2.37}{-6.59}{+1.21} & \scorecell{22.90}{29.72} & \subcell{+6.81}{-0.03}{+14.13} \\
 & Medicare & \scorecell{22.95}{25.57} & \subcell{+2.62}{-0.19}{+8.17} & \scorecell{22.98}{24.47} & \subcell{+1.49}{-1.80}{+5.60} & \scorecell{25.49}{25.70} & \subcell{+0.21}{-4.40}{+6.12} \\
 & Other & \scorecell{24.23}{23.50} & \subcell{-0.73}{-2.36}{-0.01} & \scorecell{23.52}{23.98} & \subcell{+0.46}{-0.92}{+1.99} & \scorecell{25.21}{25.35} & \subcell{+0.14}{-1.55}{+2.03} \\
\addlinespace[2pt]
Arrival mode & EMS & \scorecell{22.01}{23.96} & \subcell{+1.95}{-0.19}{+5.82} & \scorecell{22.48}{24.26} & \subcell{+1.78}{-2.45}{+6.31} & \scorecell{24.31}{29.38} & \subcell{+5.07}{-1.10}{+10.98} \\
 & Self & \scorecell{24.16}{23.55} & \subcell{-0.61}{-1.83}{-0.01} & \scorecell{23.92}{23.58} & \subcell{-0.34}{-1.58}{+0.70} & \scorecell{24.98}{24.64} & \subcell{-0.34}{-2.01}{+1.29} \\
\addlinespace[2pt]
Triage acuity & 2 & \scorecell{25.01}{22.85} & \subcell{-2.16}{-6.49}{-0.08} & \scorecell{24.13}{22.90} & \subcell{-1.23}{-3.82}{+1.13} & \scorecell{23.49}{23.18} & \subcell{-0.31}{-0.86}{+0.12} \\
 & 3 & \scorecell{23.48}{24.08} & \subcell{+0.59}{-0.07}{+1.96} & \scorecell{23.80}{24.25} & \subcell{+0.45}{-1.13}{+2.18} & \scorecell{25.47}{26.71} & \subcell{+1.24}{-1.29}{+3.82} \\
\addlinespace[2pt]
\bottomrule
\end{tabular}%
}
\end{table}

\begin{table*}[t]
\centering
\footnotesize
\setlength{\tabcolsep}{4pt}
\renewcommand{\arraystretch}{1.05}
\begin{tabular}{@{}lllccr@{}}
\toprule
\multicolumn{6}{@{}l}{\textbf{A. Detailed ordinal-axis trends}} \\
Model & Task & Axis & Trend
& \multicolumn{2}{r@{}}{Macro-F1 $\dlift$ sequence} \\
\midrule
\multicolumn{6}{@{}l}{\emph{Positive lift in all bins}} \\
QoQ-Med & \texttt{48-IHM} & CCI & inc & \multicolumn{2}{r@{}}{$\underset{\textcolor{gray}{\scriptscriptstyle[+0.5,+5.7]}}{+3.0} \to \underset{\textcolor{gray}{\scriptscriptstyle[+1.8,+6.0]}}{+4.0} \to \underset{\textcolor{gray}{\scriptscriptstyle[+5.4,+10.1]}}{+7.7} \to \underset{\textcolor{gray}{\scriptscriptstyle[+6.2,+14.1]}}{+10.1}$} \\
QoQ-Med & \texttt{48-IHM} & Age & inc & \multicolumn{2}{r@{}}{$\underset{\textcolor{gray}{\scriptscriptstyle[+0.4,+8.6]}}{+4.2} \to \underset{\textcolor{gray}{\scriptscriptstyle[+2.3,+7.3]}}{+4.8} \to \underset{\textcolor{gray}{\scriptscriptstyle[+3.3,+8.5]}}{+6.0} \to \underset{\textcolor{gray}{\scriptscriptstyle[+5.1,+10.5]}}{+7.8}$} \\
MedGemma-27B & \texttt{48-IHM} & CCI & mixed & \multicolumn{2}{r@{}}{$\underset{\textcolor{gray}{\scriptscriptstyle[+4.5,+10.1]}}{+7.4} \to \underset{\textcolor{gray}{\scriptscriptstyle[+2.8,+6.7]}}{+4.7} \to \underset{\textcolor{gray}{\scriptscriptstyle[+2.1,+6.0]}}{+4.2} \to \underset{\textcolor{gray}{\scriptscriptstyle[+2.9,+8.9]}}{+5.8}$} \\
MedGemma-27B & \texttt{48-IHM} & Age & near dec & \multicolumn{2}{r@{}}{$\underset{\textcolor{gray}{\scriptscriptstyle[+3.9,+10.3]}}{+7.0} \to \underset{\textcolor{gray}{\scriptscriptstyle[+4.1,+7.9]}}{+5.9} \to \underset{\textcolor{gray}{\scriptscriptstyle[+1.8,+6.0]}}{+4.0} \to \underset{\textcolor{gray}{\scriptscriptstyle[+3.1,+7.7]}}{+5.4}$} \\
QoQ-Med & \texttt{LOS} & CCI & mixed & \multicolumn{2}{r@{}}{$\underset{\textcolor{gray}{\scriptscriptstyle[-2.4,+5.3]}}{+1.5} \to \underset{\textcolor{gray}{\scriptscriptstyle[-1.1,+3.1]}}{+1.0} \to \underset{\textcolor{gray}{\scriptscriptstyle[-1.3,+5.3]}}{+2.0}$} \\
\addlinespace[3pt]
\multicolumn{6}{@{}l}{\emph{Negative in the lowest bin; positive in the highest bin}} \\
MedGemma-4B & \texttt{48-IHM} & Age & near inc & \multicolumn{2}{r@{}}{$\underset{\textcolor{gray}{\scriptscriptstyle[-13.3,+2.6]}}{-5.0} \to \underset{\textcolor{gray}{\scriptscriptstyle[-8.5,+1.4]}}{-3.6} \to \underset{\textcolor{gray}{\scriptscriptstyle[-2.6,+4.2]}}{+0.7} \to \underset{\textcolor{gray}{\scriptscriptstyle[-2.8,+4.1]}}{+0.5}$} \\
QoQ-Med & \texttt{ED-Revisit} & Age & near inc & \multicolumn{2}{r@{}}{$\underset{\textcolor{gray}{\scriptscriptstyle[-0.1,+0.0]}}{-0.0} \to \underset{\textcolor{gray}{\scriptscriptstyle[-4.3,-0.0]}}{-1.4} \to \underset{\textcolor{gray}{\scriptscriptstyle[-0.1,+0.2]}}{+0.0} \to \underset{\textcolor{gray}{\scriptscriptstyle[-0.4,+9.7]}}{+3.4}$} \\
QoQ-Med & \texttt{LOS} & Age & near inc & \multicolumn{2}{r@{}}{$\underset{\textcolor{gray}{\scriptscriptstyle[-9.4,+6.1]}}{-1.5} \to \underset{\textcolor{gray}{\scriptscriptstyle[-1.7,+4.6]}}{+1.4} \to \underset{\textcolor{gray}{\scriptscriptstyle[-3.8,+1.8]}}{-1.1} \to \underset{\textcolor{gray}{\scriptscriptstyle[+1.0,+5.8]}}{+3.3}$} \\
MedGemma-27B & \texttt{LOS} & CCI & mixed & \multicolumn{2}{r@{}}{$\underset{\textcolor{gray}{\scriptscriptstyle[-6.3,+0.3]}}{-3.1} \to \underset{\textcolor{gray}{\scriptscriptstyle[-5.4,-1.2]}}{-3.3} \to \underset{\textcolor{gray}{\scriptscriptstyle[-2.8,+3.0]}}{+0.1}$} \\
MedGemma-4B & \texttt{LOS} & Age & near inc & \multicolumn{2}{r@{}}{$\underset{\textcolor{gray}{\scriptscriptstyle[-4.9,-0.2]}}{-2.3} \to \underset{\textcolor{gray}{\scriptscriptstyle[-1.2,+2.5]}}{+0.6} \to \underset{\textcolor{gray}{\scriptscriptstyle[-2.2,+0.3]}}{-1.0} \to \underset{\textcolor{gray}{\scriptscriptstyle[-0.2,+2.2]}}{+0.9}$} \\
MedGemma-4B & \texttt{LOS} & CCI & inc & \multicolumn{2}{r@{}}{$\underset{\textcolor{gray}{\scriptscriptstyle[-3.1,-1.2]}}{-2.2} \to \underset{\textcolor{gray}{\scriptscriptstyle[-0.4,+1.7]}}{+0.7} \to \underset{\textcolor{gray}{\scriptscriptstyle[-0.3,+2.3]}}{+1.0}$} \\
\addlinespace[3pt]
\multicolumn{6}{@{}l}{\emph{Other patterns}} \\
MedGemma-27B & \texttt{ED-Revisit} & CCI & mixed & \multicolumn{2}{r@{}}{$\underset{\textcolor{gray}{\scriptscriptstyle[-2.3,+0.8]}}{-0.6} \to \underset{\textcolor{gray}{\scriptscriptstyle[-2.0,+6.0]}}{+1.9} \to \underset{\textcolor{gray}{\scriptscriptstyle[-1.3,+13.9]}}{+6.2} \to \underset{\textcolor{gray}{\scriptscriptstyle[-9.1,+1.2]}}{-2.7}$} \\
MedGemma-4B & \texttt{48-IHM} & CCI & mixed & \multicolumn{2}{r@{}}{$\underset{\textcolor{gray}{\scriptscriptstyle[-9.2,+1.8]}}{-3.5} \to \underset{\textcolor{gray}{\scriptscriptstyle[-2.8,+4.8]}}{+1.0} \to \underset{\textcolor{gray}{\scriptscriptstyle[-3.5,+4.1]}}{+0.2} \to \underset{\textcolor{gray}{\scriptscriptstyle[-5.0,+3.4]}}{-0.9}$} \\
MedGemma-27B & \texttt{LOS} & Age & mixed & \multicolumn{2}{r@{}}{$\underset{\textcolor{gray}{\scriptscriptstyle[-13.3,+0.7]}}{-6.0} \to \underset{\textcolor{gray}{\scriptscriptstyle[-5.2,+1.4]}}{-1.9} \to \underset{\textcolor{gray}{\scriptscriptstyle[-4.4,+0.7]}}{-2.0} \to \underset{\textcolor{gray}{\scriptscriptstyle[-5.2,-0.6]}}{-2.9}$} \\
MedGemma-4B & \texttt{ED-Decomp} & Age & near inc & \multicolumn{2}{r@{}}{$\underset{\textcolor{gray}{\scriptscriptstyle[-7.0,+0.1]}}{-2.8} \to \underset{\textcolor{gray}{\scriptscriptstyle[-2.5,+2.6]}}{-0.0} \to \underset{\textcolor{gray}{\scriptscriptstyle[-4.6,+2.1]}}{-1.0} \to \underset{\textcolor{gray}{\scriptscriptstyle[-2.6,+2.9]}}{+0.0}$} \\
MedGemma-4B & \texttt{ED-Decomp} & CCI & mixed & \multicolumn{2}{r@{}}{$\underset{\textcolor{gray}{\scriptscriptstyle[-2.1,+0.1]}}{-0.6} \to \underset{\textcolor{gray}{\scriptscriptstyle[-3.9,+3.6]}}{+0.0} \to \underset{\textcolor{gray}{\scriptscriptstyle[-6.5,+0.1]}}{-2.7} \to \underset{\textcolor{gray}{\scriptscriptstyle[-1.2,+0.3]}}{-0.4}$} \\
QoQ-Med & \texttt{ED-Revisit} & CCI & mixed & \multicolumn{2}{r@{}}{$\underset{\textcolor{gray}{\scriptscriptstyle[-0.1,+0.0]}}{-0.0} \to \underset{\textcolor{gray}{\scriptscriptstyle[-0.2,+0.1]}}{-0.0} \to \underset{\textcolor{gray}{\scriptscriptstyle[-6.5,+8.4]}}{+0.9} \to \underset{\textcolor{gray}{\scriptscriptstyle[-1.4,-0.2]}}{-0.7}$} \\
QoQ-Med & \texttt{ED-Decomp} & CCI & mixed & \multicolumn{2}{r@{}}{$\underset{\textcolor{gray}{\scriptscriptstyle[-0.1,+0.1]}}{+0.0} \to \underset{\textcolor{gray}{\scriptscriptstyle[+0.0,+0.3]}}{+0.1} \to \underset{\textcolor{gray}{\scriptscriptstyle[-0.2,+0.2]}}{+0.0} \to \underset{\textcolor{gray}{\scriptscriptstyle[+0.0,+0.7]}}{+0.3}$} \\
QoQ-Med & \texttt{ED-Decomp} & Age & mixed & \multicolumn{2}{r@{}}{$\underset{\textcolor{gray}{\scriptscriptstyle[+0.0,+0.0]}}{+0.0} \to \underset{\textcolor{gray}{\scriptscriptstyle[+0.0,+0.3]}}{+0.1} \to \underset{\textcolor{gray}{\scriptscriptstyle[-0.1,+0.3]}}{+0.1} \to \underset{\textcolor{gray}{\scriptscriptstyle[-0.1,+0.3]}}{+0.1}$} \\
\midrule
\multicolumn{6}{@{}l}{\textbf{B. Aggregate ordinal associations}} \\
Axis & \multicolumn{1}{c}{$N$} & \multicolumn{1}{c}{Mean $\rho$}
& Median $\rho$ & $\rho>0$ / $\rho<0$ & Mean high--low \\
\midrule
Charlson CCI & \multicolumn{1}{c}{12} & \multicolumn{1}{c}{$+0.18$}
& $+0.10$ & 6 / 4 & $+0.90$ \\
Age & \multicolumn{1}{c}{12} & \multicolumn{1}{c}{$+0.27$}
& $+0.60$ & 8 / 4 & $+1.66$ \\
\textbf{Pooled (CCI+age)} & \multicolumn{1}{c}{24}
& \multicolumn{1}{c}{$+0.23$} & $+0.30$ & 14 / 8 & $\mathbf{+1.28}$ \\
\midrule
\multicolumn{6}{@{}l}{\textbf{C. Variation across subgroup levels}} \\
\multicolumn{2}{@{}l}{Axis} & \multicolumn{1}{c}{$N$}
& Mean range  & Median range  & Max range  \\
\midrule
\multicolumn{2}{@{}l}{Gender (M/F)} & \multicolumn{1}{c}{12}
& $1.47$ & $1.49$ & $2.70$ \\
\multicolumn{2}{@{}l}{Age} & \multicolumn{1}{c}{12}
& $3.47$ & $3.56$ & $5.78$ \\
\multicolumn{2}{@{}l}{Charlson CCI} & \multicolumn{1}{c}{12}
& $3.65$ & $3.31$ & $8.92$ \\
\multicolumn{2}{@{}l}{Admission type} & \multicolumn{1}{c}{6}
& $8.66$ & $7.88$ & $14.04$ \\
\bottomrule
\end{tabular}
\caption{Subgroup trends and variation in macro-F1 modality lift.
\textbf{A:} $ \dlift$  sequences follow increasing age or CCI; trend labels describe
their direction. \textbf{B:} Spearman $\rho$ relates $ \dlift$  to ordinal bin index
within each model--task combination. High--low is the mean difference between
the highest- and lowest-bin lift; sign counts exclude zero correlations.
\textbf{C:} Range is max--min $ \dlift$ across subgroup levels, summarized over
$N$ model--task combinations.}
\label{tab:subgroup-trends-appendix}
\label{tab:monotonic-trends}
\label{tab:trend-aggregate}
\label{tab:axis-variation}
\end{table*}

\clearpage
\section{Attention Performance Analysis}

\begin{table}[h]
\centering
\small
\setlength{\tabcolsep}{3.5pt}
\renewcommand{\arraystretch}{1.18}
\caption{Outcome-cell ECG attention. Values are mean ECG attention percentage with 95\% bootstrap confidence intervals. $\Delta_{He-Hu}$ is ECG-helped minus ECG-hurt attention.}
\label{tab:outcome-cell-ecg-attention}
\resizebox{\textwidth}{!}{%
\begin{tabular}{@{}lllccccc@{}}
\toprule
\textbf{Dataset} & \textbf{Model} & \textbf{Task} & \textbf{Both correct} & \textbf{ECG helped} & \textbf{ECG hurt} & \textbf{Both wrong} & \textbf{$\Delta_{He-Hu}$} \\
\midrule
MIMIC & QoQ-Omni & \texttt{48-IHM}  & \subcell{37.2}{37.1}{37.3} & \subcell{38.1}{38.0}{38.2} & \subcell{37.6}{37.4}{37.8} & \subcell{38.7}{38.5}{38.9} & +0.5 \\
MIMIC & QoQ-Omni & \texttt{LOS}  & \subcell{47.1}{46.7}{47.4} & \subcell{42.7}{42.5}{43.0} & \subcell{43.8}{43.4}{44.2} & \subcell{42.3}{42.2}{42.4} & -1.0 \\
MIMIC & MedGemma-4B & \texttt{48-IHM} & \subcell{26.1}{26.0}{26.3} & \subcell{25.8}{25.4}{26.3} & \subcell{24.0}{23.7}{24.3} & \subcell{24.7}{24.2}{25.2} & +1.8 \\
MIMIC & MedGemma-4B & \texttt{LOS}  & \subcell{32.6}{32.1}{33.1} & \subcell{30.3}{29.6}{31.0} & \subcell{28.9}{28.4}{29.5} & \subcell{25.8}{25.7}{26.0} & +1.4 \\

\bottomrule
\end{tabular}
}
\end{table}

\begin{table}[h]
\centering
\small
\setlength{\tabcolsep}{3.2pt}
\renewcommand{\arraystretch}{1.16}
\caption{Selected subgroup helped-versus-hurt ECG attention gaps. Values are mean ECG attention percentages with 95\% bootstrap confidence intervals. $\Delta^{attn}_{He-Hu}$ is ECG-helped minus ECG-hurt attention.}
\label{tab:subgroup-help-hurt-attn-selected}
\resizebox{\textwidth}{!}{%
\begin{tabular}{@{}lllllccc@{}}
\toprule
\textbf{Dataset} & \textbf{Model} & \textbf{Task} & \textbf{Axis} & \textbf{Level} & \textbf{ECG helped} & \textbf{ECG hurt} & \textbf{$\Delta^{attn}_{He-Hu}$} \\
\midrule
MIMIC & MedGemma-4B & \texttt{LOS} & Admission & Observation & \subcell{37.2}{35.6}{38.7} & \subcell{32.5}{31.5}{33.5} & \subcell{+4.8}{+2.8}{+6.4} \\
MIMIC & MedGemma-4B & \texttt{LOS} & Age & 40-60 & \subcell{33.0}{31.6}{34.4} & \subcell{29.3}{28.3}{30.4} & \subcell{+3.7}{+2.0}{+5.6} \\
MIMIC & MedGemma-4B & \texttt{LOS} & Age & <40 & \subcell{34.3}{32.3}{36.2} & \subcell{28.7}{27.3}{30.4} & \subcell{+5.6}{+2.9}{+8.2} \\
MIMIC & MedGemma-4B & \texttt{LOS} & CCI & 0 & \subcell{39.4}{37.4}{41.3} & \subcell{31.2}{30.1}{32.4} & \subcell{+8.1}{+5.7}{+10.4} \\
MIMIC & MedGemma-4B & \texttt{LOS} & Insurance & Medicaid & \subcell{36.7}{34.9}{38.4} & \subcell{31.1}{29.7}{32.6} & \subcell{+5.5}{+3.2}{+7.9} \\
MIMIC & MedGemma-4B & \texttt{LOS} & Year & 2011 - 2013 & \subcell{31.6}{30.3}{32.9} & \subcell{28.3}{27.3}{29.3} & \subcell{+3.3}{+1.6}{+5.0} \\
MIMIC & QoQ-Omni & \texttt{LOS} & Admission & Observation & \subcell{47.0}{46.5}{47.5} & \subcell{50.3}{49.5}{51.1} & \subcell{-3.3}{-4.2}{-2.3} \\
MIMIC & QoQ-Omni & \texttt{LOS} & Race & ASIAN & \subcell{42.2}{40.4}{44.1} & \subcell{45.1}{42.5}{47.9} & \subcell{-2.9}{-6.1}{+0.0} \\
MIMIC & QoQ-Omni & \texttt{LOS} & Race & HISPANIC & \subcell{43.4}{42.5}{44.4} & \subcell{48.7}{47.0}{50.4} & \subcell{-5.2}{-7.2}{-3.2} \\
\bottomrule
\end{tabular}%
}
\end{table}

\end{document}